\documentclass[10pt,twocolumn,letterpaper]{article}

\usepackage{wacv}
\usepackage{times}
\usepackage{epsfig}
\usepackage{graphicx}
\usepackage{amsmath}
\usepackage{amssymb}
\usepackage{comment}

\def\wacvPaperID{14} % Enter the WACV Paper ID here

\wacvfinalcopy % *** Uncomment this line for the final submission

\ifwacvfinal
\usepackage[breaklinks=true,bookmarks=false]{hyperref}
\else
\usepackage[pagebackref=true,breaklinks=true,colorlinks,bookmarks=false]{hyperref}
\fi

\begin{document}

%%%%%%%%% TITLE
\title{Uncertainty Aware Proposal Segmentation for Unknown Object Detection}

\author{Yimeng Li and Jana Ko{\v{s}}eck{\'a}\\
George Mason University,
Fairfax, VA, USA\\
{\tt\small \{yli44, kosecka\}@gmu.edu}
% For a paper whose authors are all at the same institution,
% omit the following lines up until the closing ``}''.
% Additional authors and addresses can be added with ``\and'',
% just like the second author.
% To save space, use either the email address or home page, not both
%\and
%Second Author\\
%Institution2\\
%First line of institution2 address\\
%{\tt\small secondauthor@i2.org}
}

\maketitle
%\thispagestyle{empty}

%%%%%%%%% ABSTRACT
\begin{abstract}
Recent efforts in deploying Deep Neural Networks for object detection in real world applications, such as autonomous driving, assume that all relevant object classes have been observed during training. Quantifying the performance of these models in settings when the test data is not represented in the training set has mostly focused on pixel-level uncertainty estimation techniques of models trained for semantic segmentation. 
% It's inevitable to run into unexpected obstacles during the daily commute. A lot of them turns out to be outliers missed by regular traffic scene understanding datasets like Cityscapes. 
This paper proposes to exploit additional predictions of semantic segmentation models and quantifying its confidences, followed by classification of object hypotheses as known vs. unknown, out of distribution objects. We use object proposals generated by Region Proposal Network (RPN) and adapt distance aware uncertainty estimation of semantic segmentation using Radial Basis Functions Networks (RBFN) for class agnostic object mask prediction. The augmented object proposals are then used to train a classifier for known vs. unknown objects categories. Experimental results demonstrate that the proposed method achieves parallel performance to state of the art methods for unknown object detection and can also be used effectively for reducing object detectors' false positive rate. 
Our method is well suited for applications where prediction of non-object background categories obtained by semantic segmentation is reliable. 
% Our code is publicly available at \url{https://github.com/GMU-vision-robotics/SSeg_with_Uncertainty}
\end{abstract}

%%%%%%%%% BODY TEXT
\section{Introduction}

\begin{figure}[!t]
\begin{center}
\includegraphics[width=1\linewidth]{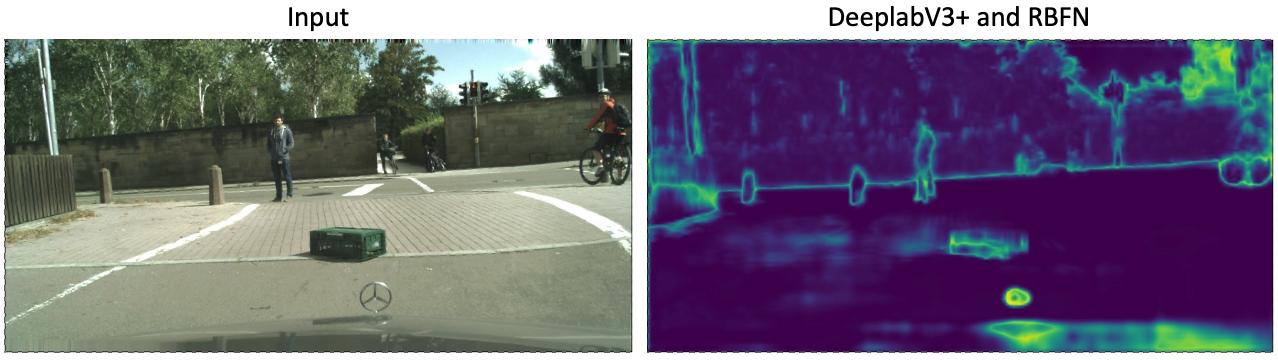}
\end{center}
   \caption{Uncertainty estimation result with DeeplabV3+ and RBFN. Note that high uncertainty sometimes show up on background regions(top right corner).}
\label{fig:dropout_prob}
\end{figure}

The last decade marked big progress in the design of Deep Network models for object detection and semantic segmentation. Highly accurate pixel-level classification of known object and background categories has been achieved by training state-of-the-art models on large fully annotated datasets~\cite{cordts2016cityscapes, lin2014microsoft}. When applying these models in real-world settings it is often the case that objects that are not represented in the training set appear at test time. This is of particular importance in autonomous driving, where unknown objects can appear on the road or close to the road and become a potential threat to safety. 
% Neural networks that use softmax normalization for confidence estimation, have % been often found assigning high confidences being assigned to out-of-distribution samples, preventing the detection of failures.
One class of methods approach this problem using methodologies for uncertainty estimation of deep networks models, such as Dropout~\cite{kendall2017uncertainties} or  ensemble methods~\cite{lakshminarayanan2017simple}.
In driving scenarios the estimation of uncertainties of semantic segmentation using Dropout often does not coincide with novel objects, making it difficult to generate reliable novel object hypotheses (see Figure.~\ref{fig:dropout_prob}). Many highly uncertain regions correspond to correctly classified background classes or boundaries between different semantic categories. 
% Second class of approaches relies on auto-encoder models, with the assumption that previously unseen objects will be decoded poorly~\cite{ganomaly2018}. 

\begin{figure}[!t]
\begin{center}
\includegraphics[width=1\linewidth]{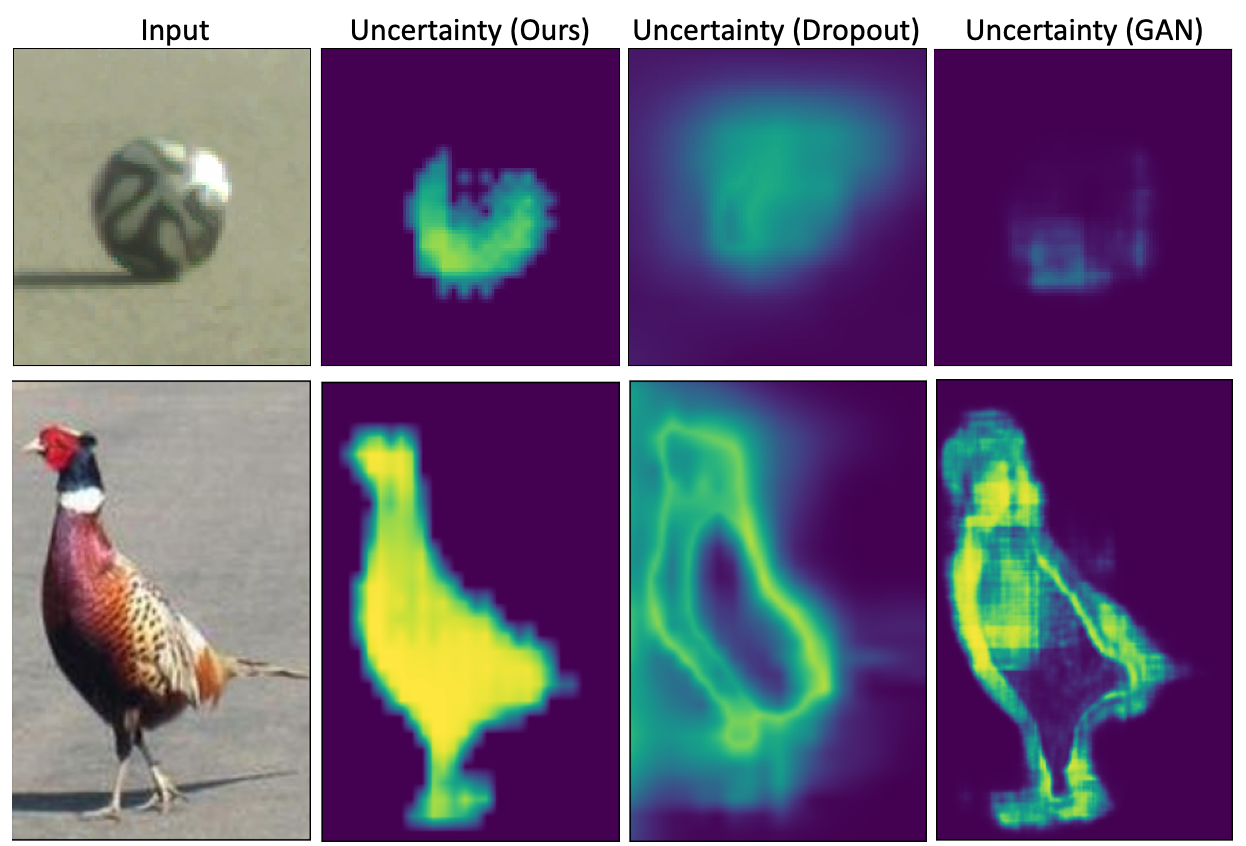}
\end{center}
   \caption{Detecting the out of distribution (OOD) objects from object proposals. Our approach predicts high uncertainty on the OOD object while Dropout~\cite{kendall2017uncertainties} method is distracted by the background and GAN~\cite{lis2019detecting} method neglects the object (football).}
\label{fig:title}
\end{figure}

% depicts our pipeline, built on the following intuition: There are always novel objects showing up in the open-set environment, but the background is stable. The pixel that certainly not belong to background categories is a candidate object pixel. 
% We apply uncertainty estimation on these possible object pixels to further pick out these confident-classified ones that belong to a known object category and the rest will be assigned to the outlier objects. 

% To overcome the difficulties of correctly labelled by highly uncertain background regions, recent approach of~\cite{lis2019detecting} used GANs to synthesize the input image from the semantic segmentation label image and trained another discrepancy network to identify the difference between the original input image and the synthesized one. While this strategy successfully handles the issues with uncertain backgrounds, it often fails when the novel objects resemble existing classes (Figure 13 of ~\cite{lis2019detecting} shows the failure case).

% Since the problem of unknown object detection required simultaneous localization and categorization of the image regions as unknown objects, the predictions made for (non-object) backgrounds classes and unknown object classes should to be treated differently. 

In this paper, we propose a new approach for detection of out of distribution objects by leveraging pixel level predictions obtained by semantic segmentation and their associated distance aware uncertainty estimates using Radial Basis Functions Networks (RBFN).
% that learn class prototypes represented by their centers in high-dimensional embedding space. 
Instead of making final predictions at the pixel level as done by methods that rely only on semantic segmentation, we use class agnostic object proposals generated by region proposal network (RPN), that are both segmented and further classified as known or unknown objects. 
The premise of our approach is that pixels belonging to background classes (\eg road, vegetation, building) can be classified with high confidence by semantic segmentation, while all object's pixels (known and unknown) will have higher uncertainty. %Once we make sure a pixel belong to an object, we can further classify this pixel into a known object category. 
%If this pixel has high uncertainty being classified into all the known object categories, then it belongs to a novel object. 
Once we make sure a pixel belong to an object, we can further decide it belongs to an unknown object if it has high uncertainty being classified into all known object categories.
This assumption is suitable for embodied agents which operate over extended periods in their native environments where non-object background classes are less likely to change.
% We propose a new approach for detection of unknown objects by leveraging existing uncertainty estimation techniques and integrating these with hypothesize-segment-classify approach, that exploits predictions made by state-of-art object detectors and semantic segmentation models. 
% We utilize the outputs of region proposal network (RPN) and show that by fine-tuning the existing semantic segmentation model for object proposals and estimating it's uncertainty with novel regularization term, 
% we can mask the highly confident pixel predictions and train a small network that classifies the object proposals into known or unknown categories. 
Our contributions can be summarized as follows:
(i) A new Radial Basis Function Network and novel regularization terms for segmentation of object proposals;
(ii) Distance aware uncertainty estimation for object mask prediction; 
%(iii) Our approach is able to improve current SSeg model's performance by fusing in object detection features in the prediction.
(iii) Detailed ablation study showing the effects of object detection and semantic segmentation features and evaluation of the proposed method on the available datasets~\cite{pinggera2016lost, blum2019fishyscapes, lis2019detecting} showing parallel performance to state-of-the-art. 
(iv) Improved the false positive rate of the modern object detector~\cite{He_2017_ICCV} and semantic segmentation models~\cite{chen2018encoder}, by identifying detections with large uncertainty.

\section{Related Work}
\noindent
{\bf Uncertainty estimation.} 
Several methods proposed for uncertainty estimation 
have been introduced in the context of image classification tasks.
In deep learning setting early approach by~\cite{boult2016} proposed an extreme value parameter redistribution method to tackle the open-set recognition problem. A similar problem to open-set recognition is to detect out-of-distribution (OOD) examples
as samples from a different dataset. Authors in~\cite{gimpel2016}
use simple statistics derived from
softmax distributions to determine whether an example is misclassified or from a different distribution from the training data.
% The presented work is People have proposed plenty of uncertainty estimation methods to evaluate the predictions made by deep neural network models. Here we review the work most relevant to our approach that applying uncertainty estimation techniques to detect out-of-dist data. 
To overcome the difficulties of stochastic methods such as dropout~\cite{kendall2017uncertainties} that require multiple network passes,
in~\cite{van2020uncertainty} propose deterministic uncertainty estimation method. They learn a feature space using RBF-kernels and suggest that feature distance to the nearest center well quantifies the uncertainty of prediction. Authors in~\cite{liu2020simple} presented a normalization method for deep neural networks to maintain the feature distance in the intermediate layers, in~\cite{li2020background} they learn a representative in-distribution data embedding by taking additional background data as adversarial examples.\\
% outlier detection on outdoor scenes
\noindent
{\bf Semantic Segmentation.} Several approaches for detecting unknown objects using pixel level predictions in semantic segmentation framework have been introduced in the autonomous driving setting along with the datasets to evaluate them.  
% ~\cite{blum2019fishyscapes, wong2020identifying, lis2019detecting}. All of them study the out-of-dist objects detection problem from the autonomous driving setting. 
Authors in~\cite{blum2019fishyscapes} proposed a dataset for this task by synthesizing unexpected objects into CityScapes images. They also implemented existing uncertainty estimation methods~\cite{gal2017concrete,Dirichlet2018} for semantic segmentation models with evaluations on the proposed dataset. The methods did not work well for detecting the unknown objects, making it difficult to aggregate pixel level uncertainty predictions to object hypotheses. This is an inherent limitation of methods based on semantic segmentation. 
Authors in~\cite{wong2020identifying} identified the outliers from Lidar-projected point cloud with prototype models used for few shot learning and reasoned about embeddings of unknown categories as distances from the prototypes. 
Instead of relying on uncertainty estimation techniques, Lis et al.~\cite{lis2019detecting} used GANs for synthesizing images from 
predicted semantic segmentation and designed a discrepancy network for identifying the differences between the original input image and a synthesized image leading the performance on available benchmarks. 
Biase et al.~\cite{di2021pixel} improve Lis's re-synthesis method by using uncertainty maps to guide the discrepancy network to focus on high-uncertain areas.
In~\cite{jung2021standardized}, authors propose to detect the unexpected objects through standardized max logits. This method does not require additional training, only requiring further processing on the segmentation results.
In~\cite{marchal2020learning}, authors use normalizing flow to learn the data density embedding space. In the test stage pixels with low density are considered as foreground objects hypotheses. The method is evaluated on indoors scenes only. \\
\noindent
{\bf Panoptic segmentation.}
The recently proposed panoptic segmentation approaches are also relevant 
to our work. Semantic segmentation model can provide pixel-level features for classification of the known object and background categories, while object detection models make predictions over bounding boxes that aggregate features over larger regions. 
Authors in~\cite{xiong2019upsnet} suggest to use both semantic segmentation features and object mask features to determine whether the pixel belongs to object or background, while authors in~\cite{weber2019single} present an instance center prediction head to assign pixels from the same class to different object instances. In~\cite{fu2019imp, hou2020real} authors show that features obtained by training object detector model help predicting higher-quality object masks compared to semantic segmentation features. In~\cite{joseph2021towards}, authors proposed a open world object detection framework that used contrastive learning in the feature space to discover unknown object classes. In presented work we also explore the efficacy of using both semantic 
segmentation features and object detection features for unknown object detection.\\
% We haven't found any existing work extending the panoptic segmentation models to novel objects.
\noindent

\section{Approach}

 We propose a novel approach for unknown object detection that starts from object proposals using RPN~\cite{He_2017_ICCV} and Edge Boxes~\cite{edge_boxes_ZitnickD14} and associated feature maps from state-of-the-art object detection and semantic segmentation models. 
 In the first stage we train a proposal segmentation model with pixel-level uncertainty for object mask prediction (see Figure~\ref{fig:title}).  
In the second stage, we pool the features from object mask region into one feature vector for  object class prediction. If a proposal has high uncertainty being classified into all known object categories, then it is labeled as an unknown object.

\begin{figure*}[!t]
\begin{center}
\includegraphics[width=0.9\linewidth]{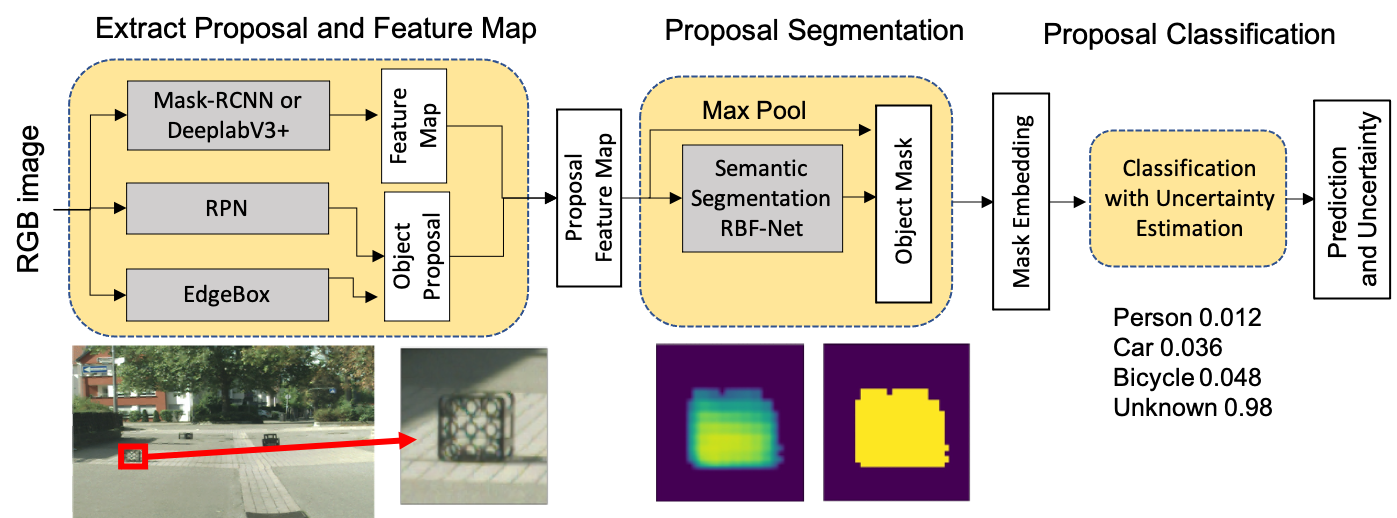}
\end{center}
%\vspace{-0.5cm}
 \caption{Proposed approach pipeline. Given an input image, we extract object proposals and proposal feature maps. Proposal segmentation module predicts the foreground object mask from the feature map. Proposal classification module pools the feature vector from the feature map given the object mask, followed by classification of object as unknown or in distribution objects and {\bf it's uncertainty estimation}. Object classification with high uncertainty is decided as an outlier.}
%\vspace{-1.5em}
\label{fig:architecture}
\end{figure*}

%***** JK Proposal Segmentation section 
\subsection{Proposal Segmentation}

The proposal segmentation model takes an object proposal $o_i$ and its associated features  $f_i$ extracted from Mask-RCNN model~\cite{He_2017_ICCV} and DeeplabV3+ semantic segmentation model~\cite{chen2018encoder}\footnote{For MASK-RCNN we take 14x14 ROI aligned feature map being passed to mask branch and for DeeplabV3 we take the feature map for all pixels in the proposal followed by ROI align stage to yield another set of channels with spatial support of 14x14.}. 
% We fine-tune {\bf JK what does fine-tune and adapt mean here ? }
% and adapt the semantic segmentation model and estimate its uncertainties for object proposals, using the original labels available in Cityscapes dataset. 
To label each pixel of the proposal into one of the semantic categories and estimate its uncertainty, we process the initial 
feature maps by adding additional convolutional layers before passing it to RBF Network. Details of the architectures are described in section~\ref{sec:implementation}.
% We are interested in learning a mapping between a pixel's feature vector $f_{i}(u,v)$ and the pixel's semantic label $c_{i}(u,v)$ and evaluate the uncertainty of the prediction.  
% {\bf ??} Particularly, we focus on the pixels on the background $c_{bg}$ so that during testing stage, pixels assigned to background with high uncertainty are potential objects $c_{obj}$, which can extend to out-of-dist objects as well. {\bf what data is this trained on} 

\noindent \textbf{RBF Network.}
For uncertainty estimation, we adapt Radial Basis Functions Network (RBFN)~\cite{abu2012learning, van2020uncertainty} and its feature-distance uncertainty estimation framework. This deterministic uncertainty quantification approach forgoes the disadvantages of dropout or ensemble methods that require multiple passes through network. The predictions of RBFN are made by computing a kernel function and a distance function, between the feature vector computed by deep model and the centroids. The uncertainty of the prediction is measured as the distance between the model output and the closest centroid.  Data points with feature vectors that are far away from centroids do not belong to any class and can be considered out of distribution. 

% by simply replacing the ResNet with a simpler feature extraction module and keeping the last RBFN classification layer.
The feature extraction module $g(f(u,v))$ consists of four convolutional layers, taking features at spatial location $f(u,v)$ in the feature map, followed by RBFN-classifier that has two learnable parts: $K$ centers $\mu_{c,k}$ for each class and a weight component $w_{c,k}$ for each center.
We apply the radial basis function (gaussian) to the feature output $g(f(u,v))$ and the class centers as a measure of distance between them:
\begin{equation}\label{distance}
    h_{c}(g(f(u,v)),\mu_c)=\sum_{k=1}^{K}w_{c,k}\exp(-\frac{\left \| g(f(u,v))-\mu_{c,k} \right \|^2}{2\sigma^{2}})
\end{equation}
where $\sigma$ is the scale term for Gaussian kernel. Class $c$ with the minimum distance (i.e. maximum $h_{c}$) is the final prediction. Uncertainty $\tau$ is computed as the difference between one (upper bound of $h_c$) and the distance to the predicted class:
\begin{equation}
    \tau=1- \max h_c(g(f(u,v)),\mu_c)
\end{equation}
The adoption of this model and uncertainty estimation framework for semantic segmentation comes with its challenges. We describe next 
how to tackle them using novel regularization term. 

\begin{figure}[!t]
\centering
\includegraphics[width=1\linewidth]{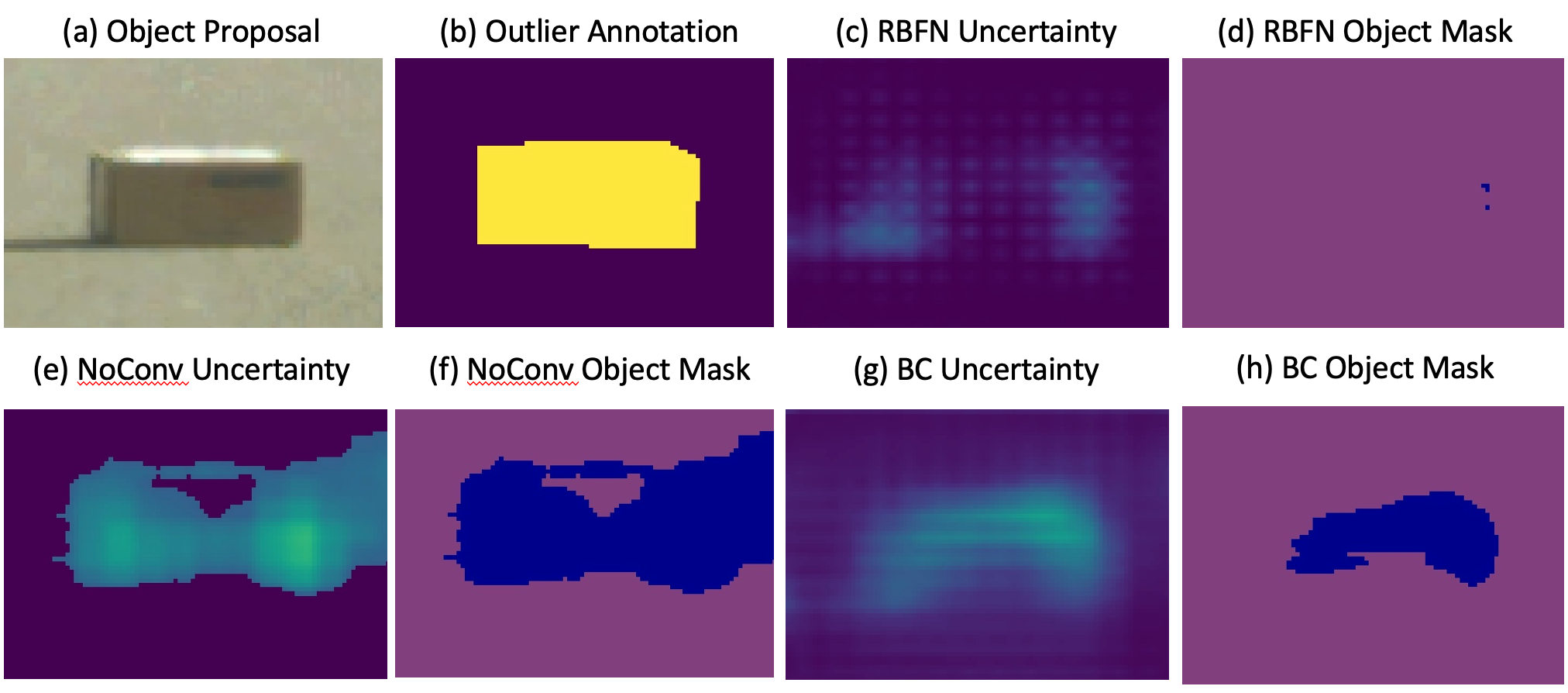}
\caption{(a)(b) show one example object proposal of out of distribution object (OOD) from Lost \& Found dataset and its annotation. (c)(d), (e)(f), (g)(h) show the proposal segmentation result using RBFN, RBFN-NoConv and RBFN with boundary constraint models. Note that enforcing the boundary constraint helps the detection of OOD object.}
\label{fig:bc_example}
\end{figure}

\noindent \textbf{Boundary Regularization.} 
In practical settings it has been observed that RBF networks are difficult to optimize and can frequently 
map the out of distribution features to in distribution features, also referred to as {\em features collapse} problem. Figure~\ref{fig:bc_example} shows an example where an out-of-distribution (OOD) object is confidently classified into the 'ground' class. 
% If we remove the convolutional layers, the model is able to separate the OOD object from the background, but suffers from generalization ability for the known classes. 
This has been observed in~\cite{van2020uncertainty}, where authors suggested adding gradient penalty to the loss function. In the context of semantic segmentation task gradient penalty is computed per pixel 
and causes loss explosion during training.  Further conflict with batch normalization, causes the gradient penalty to reduce model's overall performance. 

We propose a regularization method better suited for the segmentation task, where pixel level predictions are sought. We observed that the boundary pixels between background and object usually have high uncertainty because their receptive field includes features from both object and background pixels. If we consider these pixels as outliers, we can confine the computed embedding to be either object or background pixels. In other words, enforcing a uniform distribution for the boundary pixels $D_{bd}$ and maximizing the classification performance on the remaining in distribution pixels $D_{in}$. This is captured by the following loss function:
\begin{equation}
    L(g,w;D_{in},D_{bd})=L_{in}(g,w;D_{in})+L_{bd}(g,w;D_{bd})
\end{equation}
where
\begin{equation}
    L_{in}(g(u,v), y)=-\sum_{c} y_c \log (h_c)+(1-y_c)log(1-h_c)
\end{equation}
is the standard cross-entropy loss for in distribution pixels between each class distance $h_c$ and a one-hot encoding of the label $y$. $L_{bd}$ is the same loss for boundary pixels where the label encoding is fully zeros.
\begin{figure}[t!]
\centering
\includegraphics[width=1\linewidth]{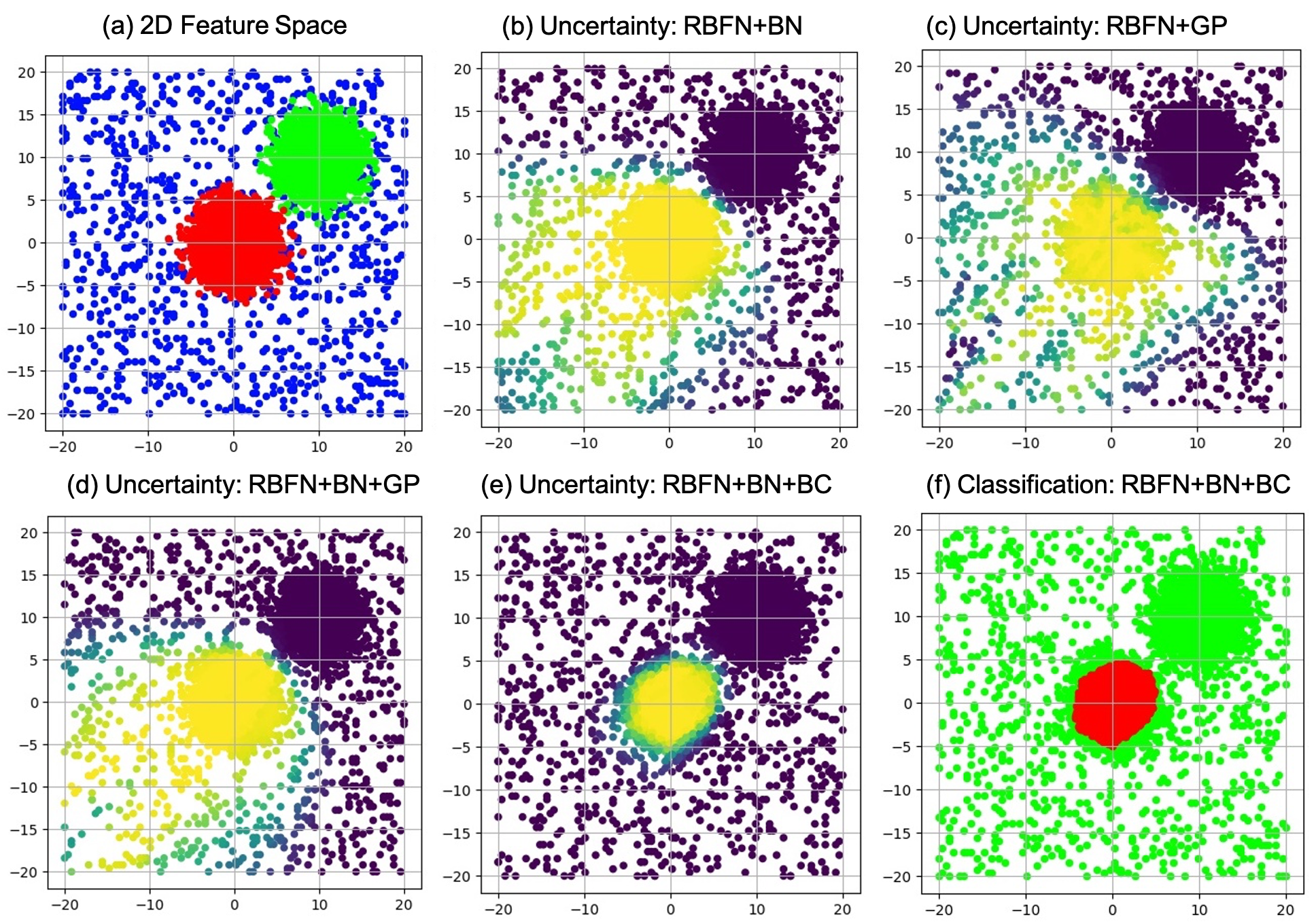}
\caption{Visualization of point classification on 2d space. (a) shows the whole feature space including two Gaussian (red Gaussian blob represents in-dist background features and green blob represents in-dist object features) and the blue points follows a uniform distribution (OOD data). (b)(c)(d)(e) shows the uncertainty estimation results (brighter color means lower uncertainty) of classifying all the points into the red Gaussian blob by using RBFN combined with different regularization methods. (f) is the final classification result by thresholding the uncertainty value and separating the red Gaussian blob from the OOD points.}
\label{fig:gaussian_example}
\end{figure}

\noindent \textbf{Toy Example in 2D Dimension.}
We use a toy example (Figure~\ref{fig:gaussian_example}) to explain the proposed regularization method.
During training stage, the in-distribution data is represented by two Gaussians, the red one for background features and the green one for object feature vectors (see  Figure~\ref{fig:gaussian_example}(a)). During testing, we add out of distribution features, here denoted by uniformly distributed blue points. We train RBF network with different regularization terms, to quantify their ability to classify out of distribution data points. 
% We use RBFN to find the OOD points: that have high uncertainty (brighter color means low uncertainty) being classified into both classes (blobs). 
Figure~\ref{fig:gaussian_example}(b) shows the estimated uncertainty of applying RBF-Net method, where some OOD points also have low uncertainty and got miss-classified into the in distribution classes. This is the feature collapse phenomenon. 
Figure~\ref{fig:gaussian_example}(c) shows that gradient penalty reduces the uncertainty distribution problem to some extent. Figure~\ref{fig:gaussian_example}(d) shows that gradient penalty contradicts with batch normalization~\cite{gouk2021regularisation} causing the number of OOD points with low uncertainty increase. The last two plots of Figure~\ref{fig:gaussian_example}(e) and (f) show that the boundary points work as a strong constraint to the point embeddings with only points in the center of the blob having high confidence (bright color). This enables us to separate the (background) points belonging to the center Gaussian from the other OOD points given the estimated uncertainty in Figure~\ref{fig:gaussian_example}(f).

The demonstration of these effects on the proposal segmentation is in Figures~\ref{fig:title} and~\ref{fig:bc_example}. The details of implementation, trade-offs between generalization ability of RBF models on known classes and ability to reliably classify out of distribution objects and the effect of different regularization terms can be found in Section~\ref{subsec:ablation_study}. 
%We also compare the proposed method with spectral normalization technique suggested by Liu~\cite{liu2020simple}. 
\begin{figure}[!t]
\centering
\includegraphics[width=1\linewidth]{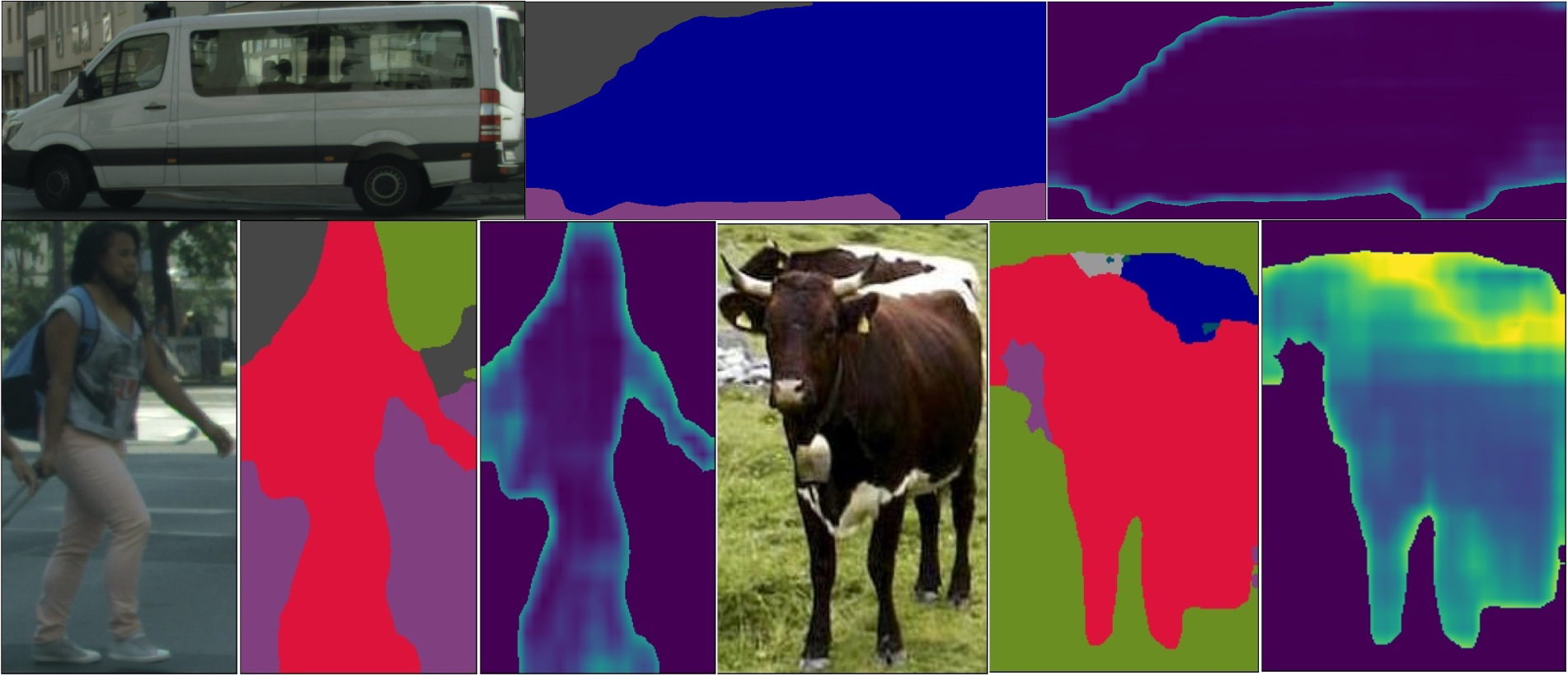}
\caption{Visualization of object proposal segmentation on Cityscapes val. Three examples are presented, including a car, a pedestrian and one OOD object. From left to right, we have the RGB input, semantic segmentation result and per pixel uncertainty estimation result. Uncertainty values on the background pixels are ignored. Note that the uncertainty values vary substantially inside the object: not only pixels inside an out-of-dist object have high uncertainty, but also pixels close to the boundary of known objects have higher uncertainty.}
\label{fig:prop_seg_prob}
\end{figure}

\subsection{Proposal Classification}
The previous sections described an approach for semantic segmentation of object proposals using uncertainty associated with pixel level predictions. 
% In natural images, semantic classes are typically defined over image regions rather than individual local pixels. In other words, a single pixel contains less information than a semantically-consistent region in images. We observe that the estimated uncertainty of proposal segmentation is unevenly distributed between the pixels belonging to the same object. This is because we lost the global information by treating each pixel independently and lead to degraded performance. In this part, we address this challenge by doing proposal classification on the proposal segmentation result.
By thresholding pixels with low uncertainty of background classes we obtain a binary object mask. We apply max-pooling on the features
associated with the mask passing the resulting feature to RBFN model for classification with uncertainty detailed in Section~\ref{sec:implementation}. We auto-label the object proposals from training data by computing the IoU between proposals and ground-truth object bounding boxes. If IoU is larger than a threshold, the proposal is labeled as the ground-truth object category. We use such labeled object proposals as training data to train the proposal classification model.
%{\bf JK how is this model trained ?}
%We observe that the uncertainty values are uneven between the same object's pixels from a proposal's semantic segmentation result. This makes sense because each pixel is evaluated independently and usually its receptive field cannot cover the whole object. However, it affects the uncertainty estimation negatively when we cannot view an object as a whole. E.g, for this cow, it has low uncertainty on most part of it because it's visually close to a 'person' object. Thus, we decide to do proposal classification on the basis of the proposal segmentation result by collecting all pixel's features.

%We classify each proposal into 5 categories: 4 object categories and 1 background category. If the uncertainty is larger than a threshold, then it will be declared as an outlier.  We first generate a general-object binary mask by taking the proposal segmentation result. All the pixels classified into object categories are recorded. Then we do max-pooling on the object pixels' features and get a feature vector of the proposal. Classification is done with the feature vector. Uncertainty is estimated either through Dropout or RBF-kernel.

\subsection{Implementation Details}
\label{sec:implementation}
 %For the proposal semantic segmentation task, the input for the proposal is 28x28 pixels and each has a 256-dim feature vector. For the proposal classification task, the input for the proposal is one 256-dim feature vector. We use the same classification module for both tasks, but vary the number of categories.
 
 %We implemented two classification module with uncertainty estimation ability. One is the dropout out module. We pass the feature vector through 2 linear layers with 1 dropout layer in the middle. The other module is the DUQ module. 
 
\noindent \textbf{Object Proposals.}
We take the top 1000 object proposals predicted by Mask-RCNN~\cite{wu2019detectron2} trained on Cityscapes. Usually 500 lower ranked proposals have large portion of background. We keep them in the training set so our model is able to distinguish between background categories. During testing, we also use object proposals generated by using EdgeBox method~\cite{edge_boxes_ZitnickD14}. \\
% Overlapped proposals with mask-rcnn proposals are filtered.
\noindent \textbf{Network Architecture Details.}
% Segmentation head
The architecture of the segmentation model is inspired by the mask head in Mask-RCNN~\cite{He_2017_ICCV}. The input feature map (14x14) is passed through 4 convolutional layers to learn the representation and one deconvolutional layer to resize the feature map from 14x14 to 28x28, before passing it to the RBFN classification module for final prediction. 
%The architecture of the classification model is the same as the segmentation model except that we apply max-pooling on the 28x28 feature map given the object mask to obtain one feature vector. 
%The segmentation model and the classification model are trained separately. 
We use the same hyperparameters for all the RBF classification layers: 512 centers for each class and feature dimension of 256-dim. The scale $\sigma$ in the Gaussian kernel is set to be 0.1.

\noindent \textbf{Model Training.}
Both proposal segmentation model and classification model is trained using SGD with initial learning rate 0.1, momentum 0.9 and batch size 64. The learning rate is divided by 10 every 10 epochs. For the RBF classification layer, we update the centers through exponential moving average method~\cite{oord2017neural} with momentum 0.999.

%\noindent \textbf{Inference and Uncertainty Estimation}

\section{Experiments}
We perform three main experiments: evaluation of the proposed model on outdoor scenes (Sec.~\ref{subsec:eval_outdoor}), ablation study of the proposed model (Sec.~\ref{subsec:ablation_study}) and evaluation of the proposed model in indoor scenes (Sec.~\ref{subsec:eval_indoor}).
To evaluate our approach on out of distribution (OOD) object detection, we compare all test methods uncertainty estimation output with the ground-truth OOD annotation and compute metrics associated with a binary classification task. We use AUROC\footnote{Area Under ROC Curve, denoted as AC in the results Table.}, to evaluate proposal segmentation performance. For evaluation on the whole image, we also compute average precision (AP) to deal with 
in-distribution and OOD data unbalanced situation. 

Since the number of false positives is also relevant for safety-critical applications, we also compute the false positive rate (FPR$_{95}$) at $95\%$ true positive rate (TPR), which is also used in~\cite{blum2019fishyscapes}. For the classification of in distribution object classes, we simply use classification accuracy (Acc).

\subsection{OOD Object Detection in Outdoor Scenes}\label{subsec:eval_outdoor}
We train the proposed model and the baseline methods on Cityscapes~\cite{cordts2016cityscapes} and evaluate on the following three datasets containing OOD objects not covered by Cityscapes.
\noindent
\textbf{FS Lost \& Found} (L\&F)~\cite{pinggera2016lost, blum2019fishyscapes}. This dataset contains 100 real images captured with the same camera setup as Cityscapes. 
Pixel-level annotations are available to distinguish between two classes, OOD objects (e.g. cargo boxes and toy cars) and classes present in Cityscapes. 
%The appearance of background classes is similar to Cityscapes and making it an ideal dataset for the evaluation of OOD object detection methods. 
We select 62 images containing objects with sufficient spatial support and object proposal size during evaluation. The unexpected objects in the rest images are neglected by EdgeBox~\cite{edge_boxes_ZitnickD14} because of unnoticeable size.\\
%Usually on one image there are from one to three OOD objects.
%The environment where these images are taken are quite different from Cityscape's environment, which brings in high difficulty on background classification. 
%Some outlier objects are quite small and close to each other, making them show up together in one proposal. 
\noindent
\textbf{Fishyscapes Static} (FS)~\cite{blum2019fishyscapes}. 
This dataset contains 30 images with unknown objects super-imposed synthetically through image compositing techniques.  Objects not covered by Cityscapes (including  aeroplane, bird, cat, cow, dog, horse, sheep) are randomly resized and positioned onto Cityscapes validation images. Postprocessing techniques like lightning and shadow adaptation are applied to make the images more genuine. \\
\noindent
\textbf{Road Anomaly} (RA)~\cite{lis2019detecting}.
This dataset contains 60 real images collected from Internet. These include OOD objects located on or near the road to mimic the traffic scenes. Various OOD objects including animals, rocks, lost tires and construction equipment are present. Note that most images from this dataset have a very different background setting than Cityscapes. We evaluate on this dataset to compare generalization of different methods to other outdoor scenes.

\noindent \textbf{Baselines.}
We switch DeeplabV3+ last classification layer with RBFN~\cite{van2020uncertainty} as the first baseline and denote it as DeeplabV3+-RBFN. Pixel-level uncertainty is computed as per pixel's feature distance to the closest class centers. 

We use the GAN method~\cite{lis2019detecting} as the second baseline. It uses CycleGAN to generate a synthetic image from the semantic segmentation of the input image. The synthesis on the OOD object regions is expected to be poor as these outliers are not covered in the training data. We take the pixel discrepancy image between input image and synthesized image as the uncertainty estimation result for comparison since the discrepancy value is in the range $[0,1]$. 

We also compare to the state of the art method~\cite{di2021pixel} that came out a few months ago and denote it as Resynthesis++. This method is built on th GAN method\cite{lis2019detecting} while the uncertainty maps are also considered in the final inference. We take the trained model and evaluate it for the Whole Image Segmentation task only.

\noindent \textbf{Proposal Segmentation.}
We compare our proposal segmentation method with the baseline methods on proposal segmentation task. 
Object proposals are chosen if they overlap with any OOD objects. We treat proposal segmentation on OOD objects as a binary segmentation task, where one pixel's uncertainty of classification indicates the probability of it belonging to an OOD object. Proposal segmentation results for DeeplabV3+-RBFN and GAN are cut out from the whole image uncertainty map results. The proposal segmentation model uses semantic segmentation (SSeg) features from pre-trained DeepLabV3+ model as input. Table~\ref{table:prop_seg} presents the result. Our method performs particularly well on L\&F dataset. This is because images in L\&F have similar background to the training data. On FS, our model performs slightly worse than the GAN method. We hypothesise that since synthetic OOD objects in FS are blended into the background, they have a small distance to background in the feature space, which is detrimental to the feature distance based methods.
\begin{table*}[ht]
\begin{center}
\begin{tabular}{|c |c|c|c|}
\hline
Method & \shortstack{L\&F \\ (AC$\uparrow$/AP$\uparrow$/FPR$_{95}\downarrow$)} & \shortstack{RA \\ (AC$\uparrow$/AP$\uparrow$/FPR$_{95}\downarrow$)} & \shortstack{FS \\ (AC$\uparrow$/AP$\uparrow$/FPR$_{95}\downarrow$)} \\
\hline
\hline
DeeplabV3+-RBFN~\cite{van2020uncertainty} & 73.8 / 40.3 / 55.5 & 60.1 / 39.9 / 72.3 &  82.7 / \textbf{64.3} / 42.8 \\
\hline
GAN~\cite{lis2019detecting} & 85.8 / 58.5 / 33.1 &  70.6 / 54.0 / 55.7 & \textbf{84.0} / 63.9 / \textbf{40.0} \\
\hline
Ours & \textbf{92.1} / \textbf{70.3} / \textbf{23.4} & \textbf{76.2} / \textbf{56.5} / \textbf{47.3} & 82.8 / 56.3 / 43.0\\
\hline
\end{tabular}
\end{center}
\caption{Comparison of Proposal Segmentation Performance.}
\label{table:prop_seg}
\end{table*}

%\noindent \textbf{Proposal Classification.}
%Here we evaluate the performance of the classification model. Evaluation is done on Cityscapes val and the three outlier datasets. We compute the classification accuracy on the 1000 proposals of each Cityscapes validation image. For the out-of-dist object proposals from the outlier images, we check if the estimated uncertainty is above a certain threshold. We also evaluate the uncertainty on Cityscapes validation proposals to see if they all have a low uncertainty.

\begin{figure*}[!t]
\begin{center}
\includegraphics[width=0.8\linewidth]{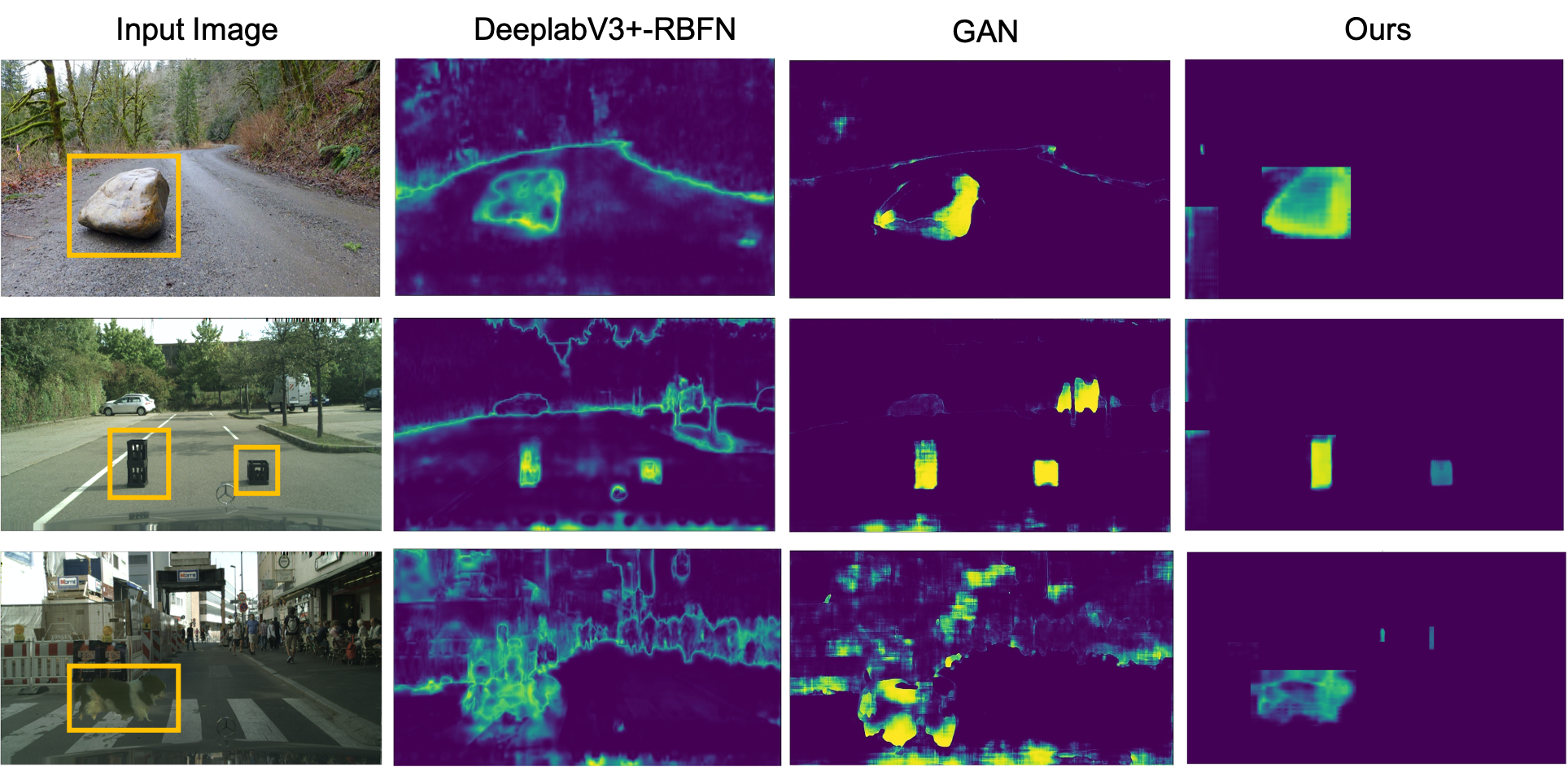}
\end{center}
%\vspace{-0.5cm}
 \caption{Whole Image OOD Object Segmentation Results. Orange box denotes the OOD object. Each row corresponds to a different input image. From top to bottom, the input image is selected from Road Anomaly, Lost \& Found and Fishyscapes datasets. From left to right, we have the input image, Deeplab-RBFN~\cite{van2020uncertainty} result, GAN~\cite{lis2019detecting} result and our method result.}
%\vspace{-1.5em}
\label{fig:whole_img_seg}
\end{figure*}

\noindent \textbf{Whole Image Segmentation.}
Here we compare the performance of our method with the baselines on the entire image. We first rank object proposals by their objectness score, removing the ones with large IoU.
Proposal's feature map is then passed through the proposal segmentation (Prop-Seg) and proposal classification (Prop-Cls) model to compute per pixel uncertainty $u_{seg}$ and proposal's overall uncertainty $u_{cls}$. Proposal segmentation uses SSeg features as input while Prop-Cls uses object detection (ObjDet) features obtained from training Mask-RCNN on Cityscapes. If $u_{cls}$ is below a threshold (in practice, we use 0.3), then this proposal's result is discarded, otherwise per pixel uncertainty is computed as $u_{seg} \cdot u_{cls}$. We accumulate the results from all the remaining proposals and embed them to an empty uncertainty image as the final result. Table~\ref{table:whole_seg} presents the results. Our method achieves parallel performance to the recent Resynthesis++ method across all the datasets. It performs quite well on AP compared to baselines as it is less affected by the uncertainty computed on the background regions. It performs slightly worse on Road Anomaly dataset as its images are collected from Internet and the background is different from images in Cityscapes. This results in our method mistakenly recognizing some background proposals as OOD objects. Figure~\ref{fig:whole_img_seg} shows some qualitative results of the proposed method and baselines.

\begin{table*}[ht]
\begin{center}
\begin{tabular}{|c |c|c|c|}
\hline
Method & \shortstack{L\&F \\ (AC$\uparrow$/AP$\uparrow$/FPR$_{95}\downarrow$)} & \shortstack{RA \\ (AC$\uparrow$/AP$\uparrow$/FPR$_{95}\downarrow$)} & \shortstack{FS \\ (AC$\uparrow$/AP$\uparrow$/FPR$_{95}\downarrow$)} \\
\hline
\hline
DeeplabV3+-RBFN~\cite{van2020uncertainty} & 68.9 / 3.3 / 54.7  & 73.2 / 20.0 / 54.1  & 78.2 / 14.7 / 44.9 \\
\hline
GAN~\cite{lis2019detecting} & 84.2 / 10.1 / 28.9 & \textbf{86.1} / 42.3 / \textit{32.2} & 82.6 / 16.1 / \textit{40.2} \\
\hline
Resynthesis++~\cite{di2021pixel} & \textbf{95.2} / \textbf{53.8} / \textbf{13.8} & 84.6 / 41.5 / \textbf{45.6} & \textbf{92.7} / \textbf{56.3} / \textbf{26.5} \\
\hline
Ours & \textit{90.7} / \textit{32.8} / \textit{24.9} & 78.7 / \textbf{45.3} / 37.3 & \textit{88.0} / \textit{34.3} / 41.5 \\
\hline
\end{tabular}
\end{center}
\caption{Whole Image Segmentation Performance. Bold numbers denote the top performance and italic numbers denote the second performance.}
\label{table:whole_seg}
\end{table*}

\subsection{Ablation Study}\label{subsec:ablation_study}
In this section we attempt to get a better understanding of how certain components of our model contribute to the overall performance. We evaluate four aspects of the proposal segmentation (Prop-Seg) model: input representation, uncertainty estimation method, training data and regularization method. The proposal classification (Prop-Cls) model is affected by two aspects: input representation and training data. All the evaluations are done over object proposals overlapping with the OOD object labels.

\noindent \textbf{Feature Representation: ObjDet vs. SSeg}
We vary the input visual representation with remaining aspects fixed. Results for proposal segmentation model (Prop-Seg) are presented in lines 1, 3 of Table~\ref{table:ab_study_seg}, and lines 1, 3 of Table~\ref{table:ab_study_cls} for proposal classification (Prop-Cls). For the proposal segmentation task, using SSeg feature performs better than ObjDet feature particularly with the FPR$_{95}$ metric under all three datasets. This is because the SSeg feature is optimized over the background pixels while ObjDet feature is mainly trained with the object pixels. This endows SSeg features with a better description power in distinguishing the foreground objects from the background. For the proposal classification task, ObjDet features beat SSeg features by over $30\%$ AP on the OOD datasets. This shows that ObjDet features are more suitable even for the OOD objects.

\noindent \textbf{Regularization}
We compare the regularization methods for Prop-Seg. Results are presented in line 7 (no convolutional (no-conv) layers so no need for regularization), line 3 (original model), line 2 with gradient penalty (GP) and line 8 with our proposed boundary pixel constraint of Table~\ref{table:ab_study_seg}. Note that the no-conv model performs well on FS but not so much on L\&F and RA. This is because in FS the OOD objects are synthesized onto the image while in L\&F and RA we have real OOD objects. Without having the convolutional layers to further fine-tune the input data for the RBFN layer, the no-conv model generalizes poorly to new data. This is denoted by the high FPR$_{95}$ on L\&F. 

Comparing line 1 with line 2 of Table~\ref{table:ab_study_seg}, we observe that adding gradient penalty (GP) during training hinders the model's performance. We don't show the model using SSeg feature as input and having GP for regularization because we encountered loss explosion when training the model. On the other hand, the model trained with the boundary pixel constraint performs slightly better on AC and FPR$_{95}$. This means that with the help of the boundary constraint, our model learns a more robust representation and is able to detect hard OOD object examples as showed in Figure~\ref{fig:bc_example}. This is important for safety-critical applications.

\noindent \textbf{Uncertainty Estimation Methods}
Here we compare different uncertainty estimation methods. We experiment with three techniques: Entropy, Dropout and RBFN. The entropy method is implemented by using a linear layer in the end of the model for classification. Uncertainty is computed as the entropy of the output probabilities. The Dropout method is implemented by adding dropout layers after all convolutional layers. Uncertainty is estimated by performing multiple forward passes through the model with dropout enabled, and computing the entropy of the averaged predicted probability vector. Line 3, 4 and 5 of Table~\ref{table:ab_study_seg} present the results. RBFN outperforms the Entropy and Dropout by a large gap on L\&F and RA. However, none of the three methods perform well on FS. This indicates that current uncertainty estimation methods are not sensitive to synthesized outliers.

\noindent \textbf{Training Data}
Here we evaluate how the training data affects the Prop-Seg and Prop-Cls model's performance. Marshal~\cite{marchal2020learning} suggests to perform density estimation with background features only for the background segmentation task. We followed their idea and trained a Prop-Seg model using only the background pixels (Line 6 of Table~\ref{table:ab_study_seg}). Comparing with line 3 where all class labels are being used, model trained with only background pixels performs much worse across all datasets. This shows the effectiveness of having negative examples (object pixels) during training even for feature distance based models.
\begin{comment}
%Even though the task is foreground object segmentation, apparently the feature is more representative when the model is trained with all classes so that it can better distinguish object between background.
For proposal classification model (Prop-Cls), the granularity of class labels affects the performance. We trained two models with exactly the same input data but vary the class labels. Cityscapes contains 9 object labels including 'poles', 'person', 'car', 'rider', 'truck', 'bus', 'bicycle', 'motorcycle' and 'train'. % Initially to reduce the distraction of uncertainty between in-dist classes, we merge visually similar in-dist classes and 
We reduce the number of classes to 4: merge 'rider', 'bicycle', 'motorcycle' with 'person', 'bus', 'truck' with 'car'. Line 1 and 2 of Table~\ref{table:ab_study_cls} present the results. The results coincide with the recent finding in~\cite{joseph2021towards} that having labels of finer categories is beneficial to learning more suitable feature representations, in this case improving classification of OOD objects by over $10\%$AP.

% In contrary to our intuition, having finer class labels improves the classification model's performance on OOD objects by over $10\%$AP.
\end{comment}

\begin{table*}[ht]
\begin{center}
\begin{tabular}{|c|c|c|c|c|c|c|c|}
\hline
r & \shortstack{Input \\ Rep} & \shortstack{Uncertainty \\ Estimation}  & \shortstack{Trained \\ Classes}  & Regularization & \shortstack{L\&F \\ (AC$\uparrow$/AP$\uparrow$/FPR$_{95}\downarrow$)} & \shortstack{RA \\ (AC$\uparrow$/AP$\uparrow$/FPR$_{95}\downarrow$)} & \shortstack{FS \\ (AC$\uparrow$/AP$\uparrow$/FPR$_{95}\downarrow$)}\\
\hline
\hline
1 & ObjDet & RBFN & All & - & 84.8 / 59.2 / 37.9 & 76.1 / \textbf{60.7} / 49.5 & 74.4 / 48.6 / 48.9\\
\hline
2 & ObjDet & RBFN & All & \shortstack{Gradient \\ Penalty} & 84.0 / 61.5 / 43.1 & 73.3 / 57.5 / 55.4 & 68.1 / 43.9 / 61.0\\
\hline
3 & SSeg & RBFN & All & - & 90.4 / \textbf{71.7} / 30.9 & 74.0 / 59.5 / 48.0 & 81.0 / 61.3 / 44.9 \\
\hline
4 & SSeg & Dropout & All & - & 81.6 / 52.7 / 45.6 & 70.6 / 50.0 / 62.8 & 82.4 / 59.8 / 42.7\\
\hline
5 & SSeg & Entropy & All & - & 79.3 / 49.4 / 49.7 & 71.9 / 53.8 / 61.4 & 77.6 / 53.4 / 51.3\\
\hline
6 & SSeg & RBFN & bg only & - & 74.1 / 43.9 / 21.3 & 68.0 / 48.6 / 62.8 & 69.7 / 49.3 / 62.6\\
\hline
7 & SSeg & RBFN & All & No Conv & 82.4 / 56.2 / 43.4 & 74.7 / 60.4 / 52.0 & \textbf{83.3} / \textbf{62.0} / \textbf{40.2} \\
\hline
8 & SSeg & RBFN & All & \shortstack{Boundary \\ Constraint} & \textbf{92.1} / 70.3 / \textbf{23.4} & \textbf{76.2} / 56.5 / \textbf{47.3} & 82.8 / 56.3 / 43.0\\
\hline
\end{tabular}
\end{center}
\caption{Ablation Study on our Proposal Segmentation Model, illustrating the performance of models trained with different visual input, different uncertainty estimation techniques and without the regularization methods.}
\label{table:ab_study_seg}
\end{table*}

%\noindent \textbf{Input Representation for Proposal Classification}
\begin{table}[ht]
\begin{center}
\begin{tabular}{|c|c|c|c|}
\hline
Method & \shortstack{Cityscapes \\ (Acc)} & \shortstack{L\&F \\ (AC/AP)} & \shortstack{RA \\ (AC/AP)} \\
\hline
\hline
ObjDet (9 classes) & \textbf{97.8} & \textbf{95.5} / \textbf{40.5} &  \textbf{95.4} / \textbf{67.1} \\
\hline
ObjDet (4 classes) & 97.1 & 93.3 / 25.3 &  88.3 / 44.5 \\
\hline
SSeg (9 classes) & 93.1 & 80.3 / 8.7  & 78.5 / 28.2\\
\hline
SSeg (4 classes) & 89.2 & 86.1 / 11.5  & 68.4 / 17.1\\
\hline
\end{tabular}
\end{center}
\caption{Proposal Classification on Outdoor Scenes.}
\label{table:ab_study_cls}
\end{table}

\subsection{Mask-RCNN false positives}\label{subsec:maskrcnn}

\begin{figure}[ht]
\begin{center}
\includegraphics[width=1.0\linewidth]{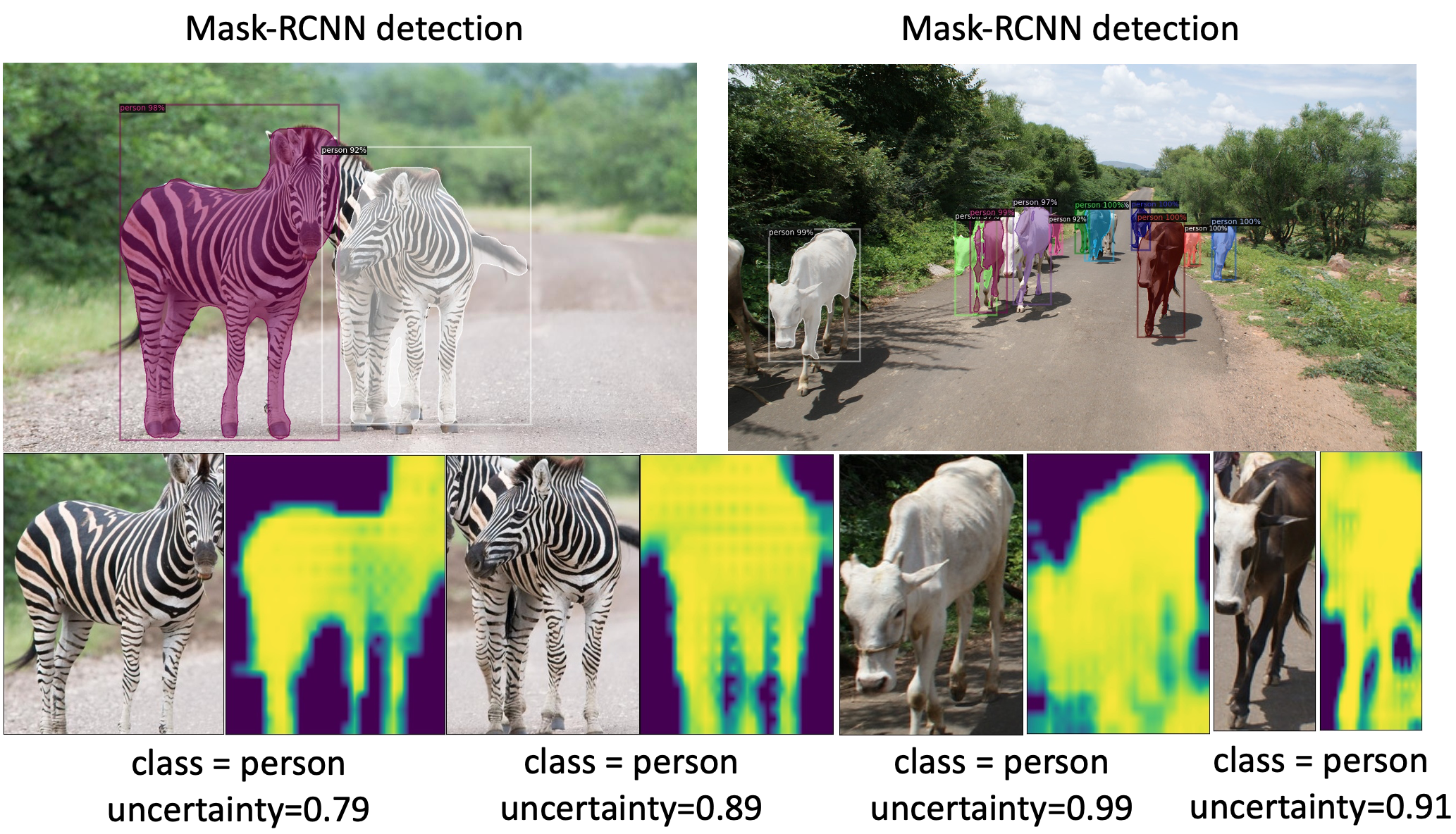}
\end{center}
   \caption{Compare Mask-RCNN predictions with our model. Mask-RCNN detects the OOD animals as 'person' with high score (over $95\%$). While our model also classifies these animals into 'person' but with high uncertainty.}
\label{fig:maskrcnn_prob}
\end{figure}

% Mask-RCNN sometimes detects OOD objects as in-dist objects with high confidence. 
We take the proposals and their feature maps from Mask-RCNN trained on Cityscapes and pass them into our proposal classification model. The model miss-classifies the same objects, but correctly estimates the high uncertainty of the predictions.  Figure~\ref{fig:maskrcnn_prob} shows two examples from Road Anomaly dataset where animals are confidently predicted as 'person' by Mask-RCNN. 
% The effect of uncertainty estimates of our model is summarized in Figure~\ref{fig:hist_fasle_detection}. 
There are 13 wrong detections and 12 of them have uncertainty above $0.5$.

\subsection{OOD Object Detection in Indoor Scenes}\label{subsec:eval_indoor}
\begin{figure}[ht]
\begin{center}
\includegraphics[width=1\linewidth]{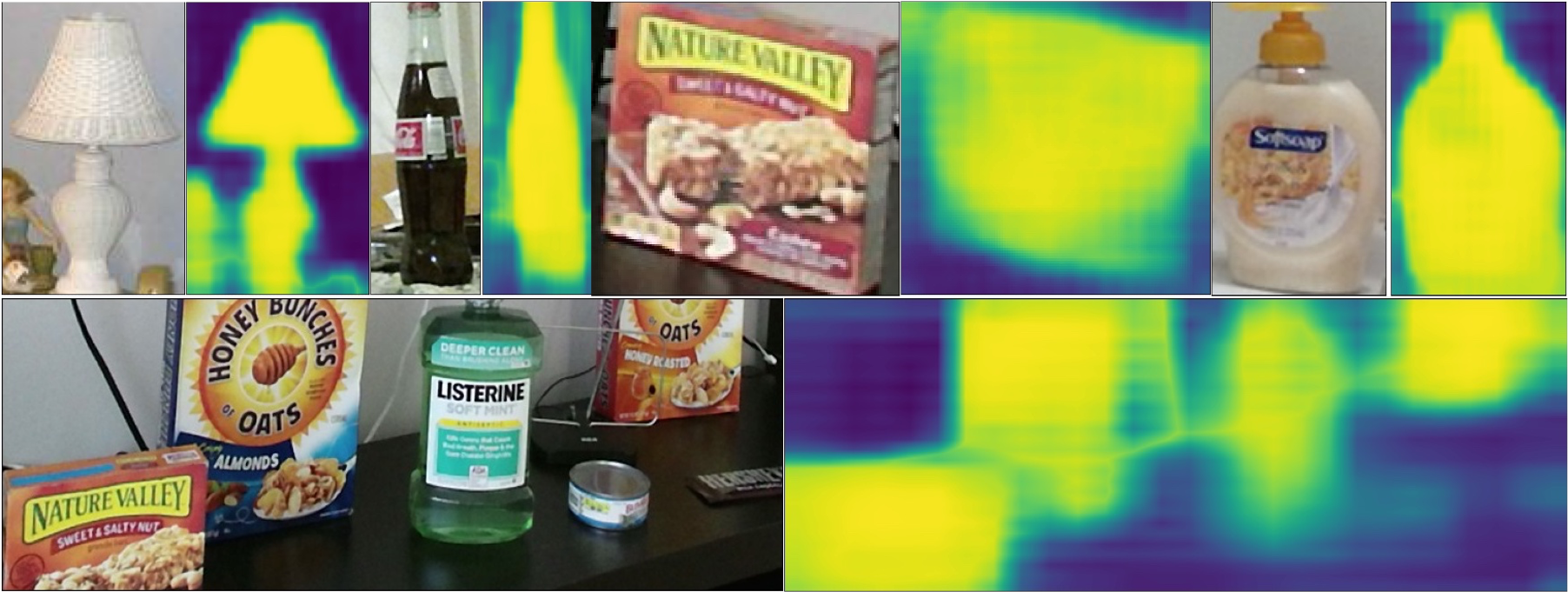}
\end{center}
   \caption{Top: segmentation of novel ADE20K and AVD instances. Bottom: uncertainty based segmentation, for proposals with  more than one object. Note that the uncertainty on the table and wall is apparently lower than the foreground unknown objects.}
\label{fig:indoor_seg}
\end{figure}

We train a Mask-RCNN and DeeplabV3+ on ADE20K~\cite{zhou2017scene} and extract proposals and ObjDet/SSeg feature maps. Two object classes, 'vase' and 'lamp' are ignored during training. During testing, we select object proposals containing these two classes and evaluate our approach 
on ADE20K and  AVD~\cite{ammirato2017dataset} datasets. The results with different input representation are in Table~\ref{table:prop_seg_indoor}. The model using ObjDet features performs slightly better. In indoor scenes proposals have fewer background pixels, taking out the advantage of SSeg features for background representation. Figure~\ref{fig:indoor_seg} shows proposal segmentation results.

\begin{table}[ht]
\begin{center}
\begin{tabular}{|c|c|c|}
\hline
Method & \shortstack{ADE20K \\ (AC$\uparrow$/AP$\uparrow$/FPR$_{95}\downarrow$)} & \shortstack{AVD \\ (AC$\uparrow$/AP$\uparrow$/FPR$_{95}\downarrow$)} \\
\hline
\hline
ObjDet  & \textbf{92.3} / \textbf{92.8} / \textbf{26.8} &  \textbf{94.9} / \textbf{97.0} /  17.9 \\
\hline
SSeg  &  90.9 / 89.4 / 31.3 & 94.3 / 96.4 /\textbf{16.8} \\
\hline
\end{tabular}
\end{center}
\caption{Proposal Segmentation on Indoor Scenes: ADE20K and AVD.}
\label{table:prop_seg_indoor}
\end{table}

%\subsection{Improve Semantic Segmentation}

%\subsection{Proposal Segmentation with Other Features}
%Here we are trying to evaluate if a general object detector's features can generalize to out-of-dist objects or not. We extract proposal features from a Mask-RCNN model trained on LVIS dataset and train a new proposal segmentation model. It exhibits a parallel performance to the model using Cityscapes Object Detector's features. This experiment tells us that an open-set object detector is promising to train with a closed-set object detector.

\section{Conclusion}
We proposed a two-step proposal segmentation and classification method using RBFN for unknown object detection. We examine the performance of the proposal segmentation model using different backbone features and a variety of regularization methods. The proposed regularization through boundary pixel constraint proved to be most useful for finding hard out-of-distribution examples. We present comprehensive comparison of the model to alternative approaches in the literature. The proposed method can be used to flag false positives made by modern object detectors. In the experiment, we also demonstrate the method's generalization to indoor scenes. In the future we plan to integrate the RBFN prototype model into a region proposal network to detect general objects more effectively. We are also interested in seeing if the proposed method can detect adversarial attacks on modern semantic segmentation and object detection models. 

%\clearpage
{\small
\bibliographystyle{ieee_fullname}
\bibliography{egbib}
}

\clearpage

\section{Supplementary Material}

\subsection{Additional Details of Sec. 4.4}
We select 4000 training and 550 testing images from ADE20K Scene Parsing dataset from the following indoor scene types: 'bathroom', 'bedroom', 'kitchen', 'living room', 'office', 'dining room', 'hotel room', 'dorm room', 'home office', 'waiting room'.

The proposal segmentation model is trained with these known classes: wall, floor, ceiling, bed, window, cabinet, door, table, plant, curtain, chair, painting, sofa, shelf, mirror, carpet, bathtub, cushion, sink, fridge, toilet. And \textit{lamp} is the novel class we left out to detect during testing.

\subsection{Additional Results}
\begin{figure*}[ht]
\begin{center}
\includegraphics[width=0.8\linewidth]{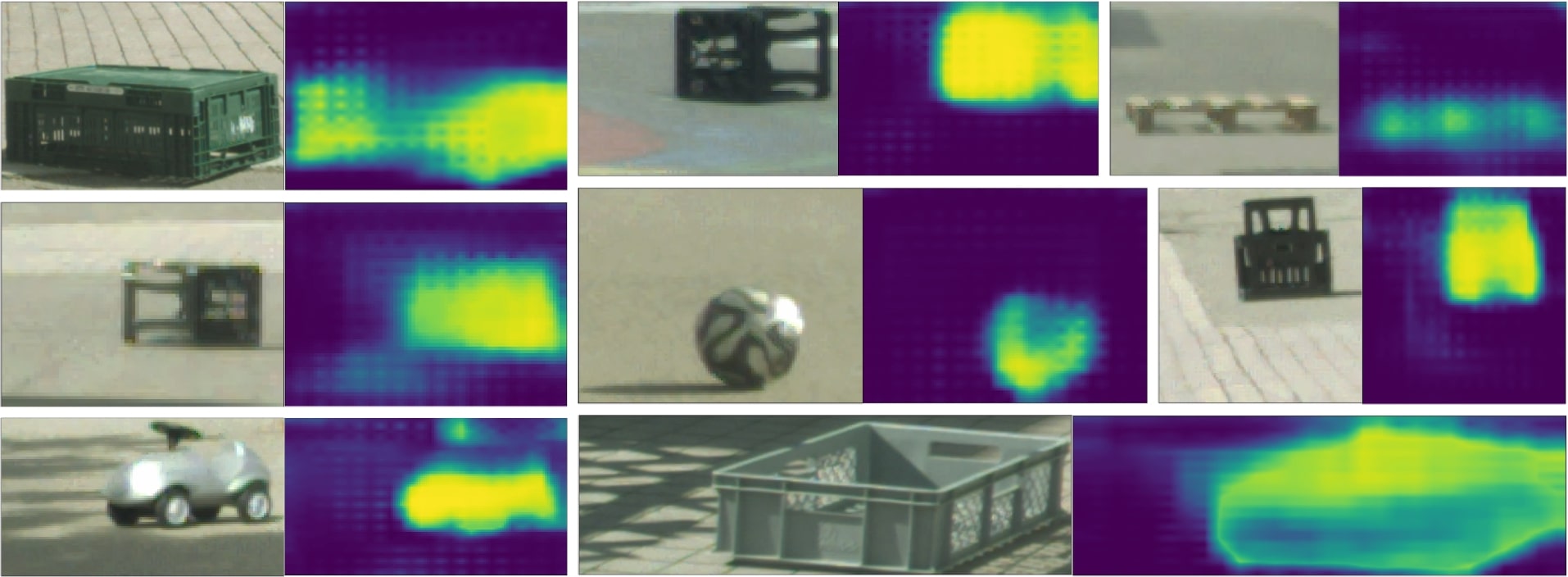}
\includegraphics[width=0.8\linewidth]{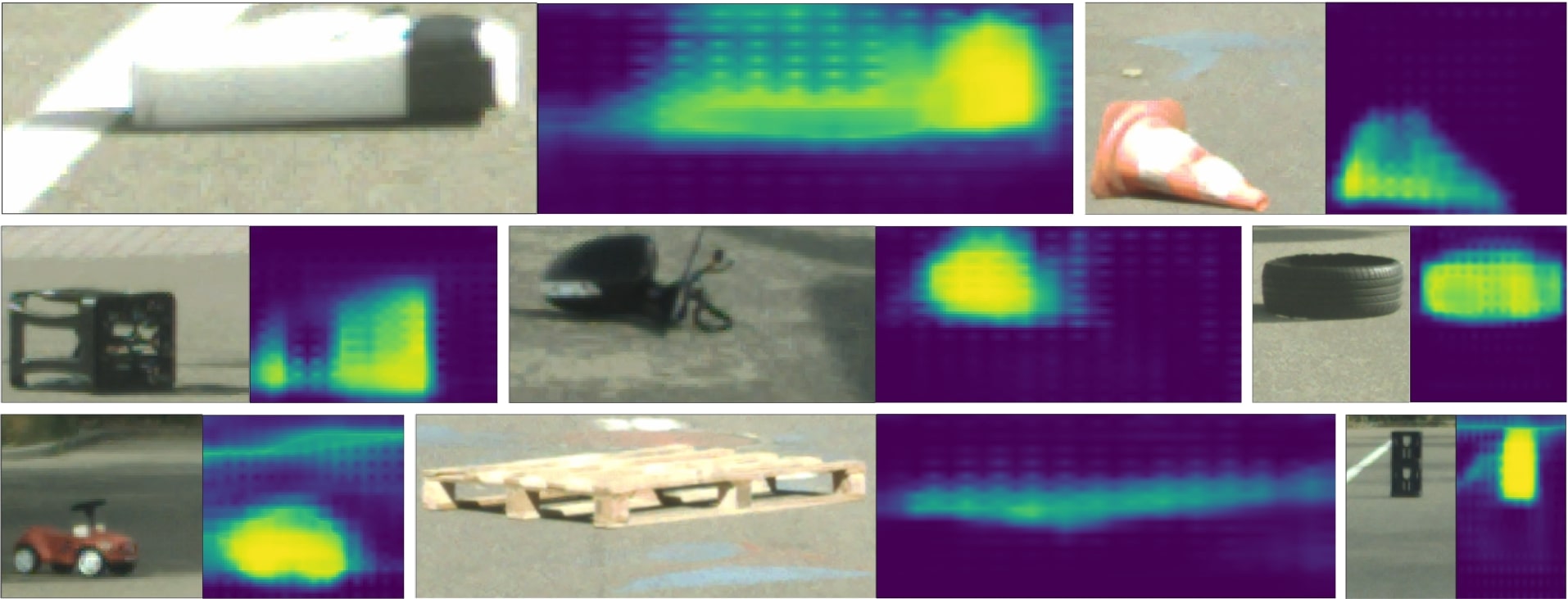}
\includegraphics[width=0.8\linewidth]{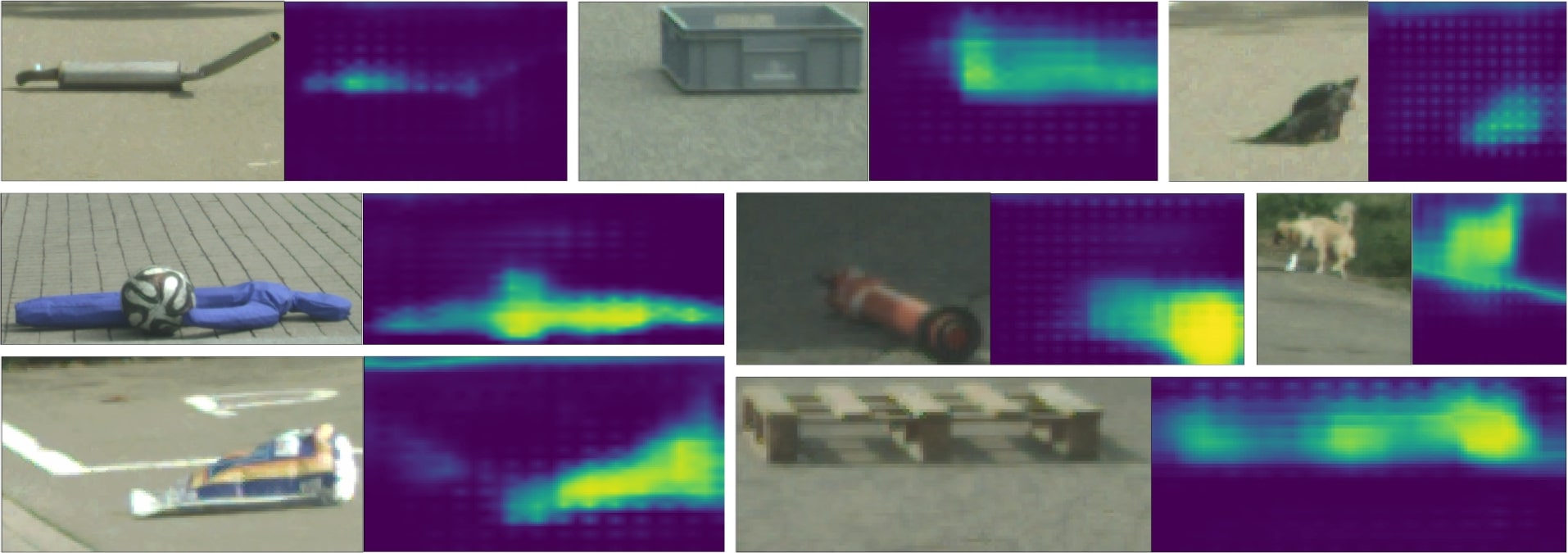}
\includegraphics[width=0.8\linewidth]{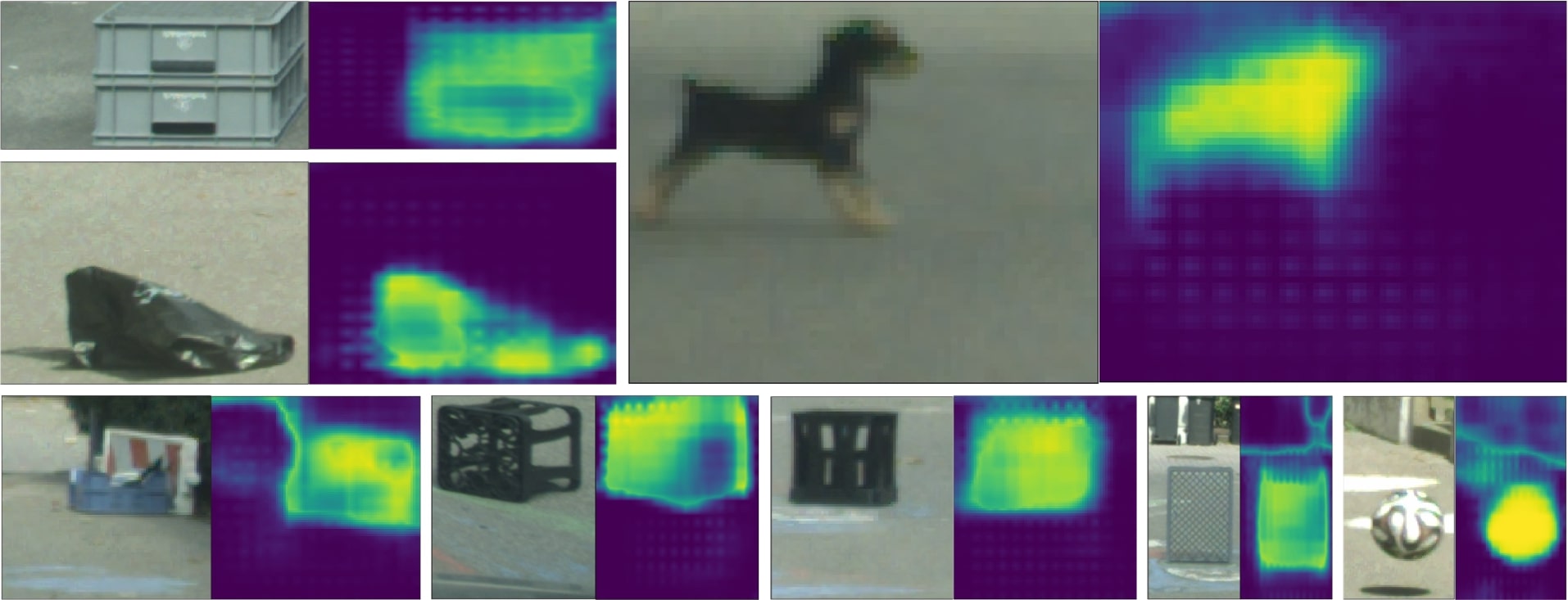}
\end{center}
 \caption{OOD proposal segmentation on Lost\&Found dataset. We showed the input proposal and uncertainty estimation results.}
\label{fig:prop_seg}
\end{figure*}

\begin{figure*}[ht]
\begin{center}
\includegraphics[width=0.9\linewidth]{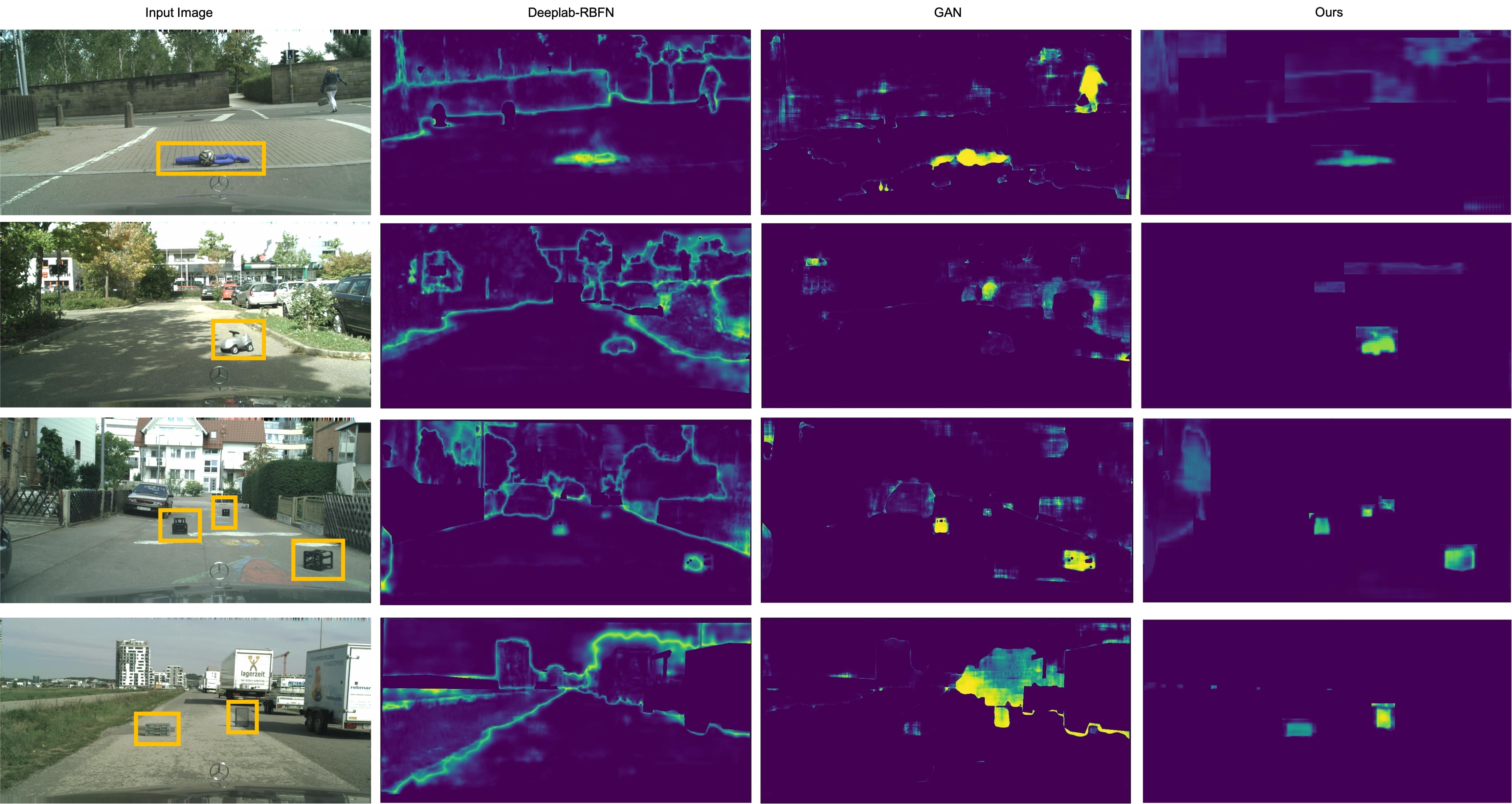}
\end{center}
 \caption{Whole image OOD object segmentation results on Lost\&Found dataset. From left to right are the input image, Deeplab-RBFN method result, GAN method result and our method result.}
\label{fig:whole_LF}
\end{figure*}

\begin{figure*}[ht]
\begin{center}
\includegraphics[width=0.9\linewidth]{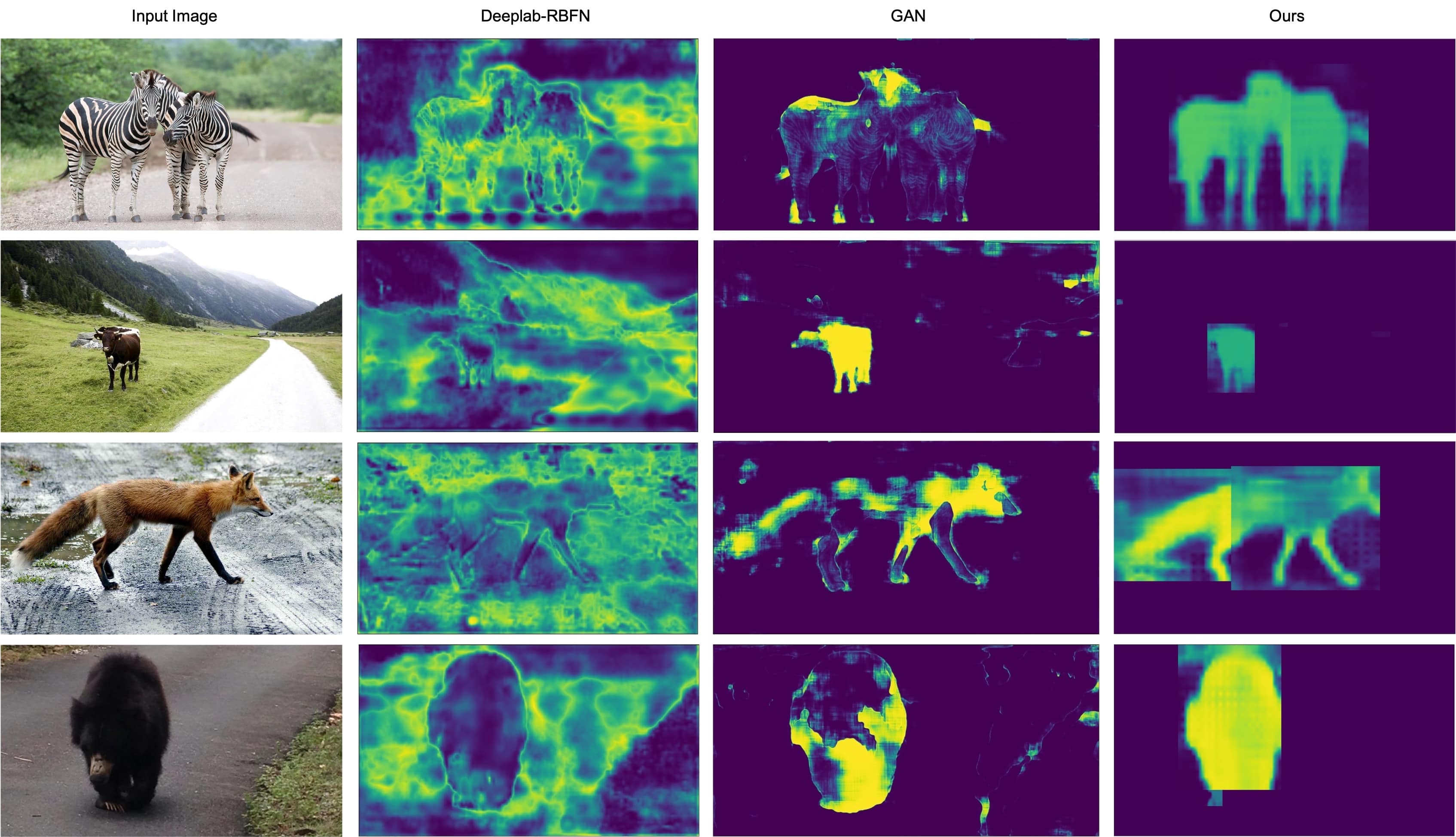}
\end{center}
 \caption{Whole image OOD object segmentation results on RoadAnomaly dataset. From left to right are the input image, Deeplab-RBFN method result, GAN method result and our method result.}
\label{fig:whole_RA}
\end{figure*}

\begin{figure*}[ht]
\begin{center}
\includegraphics[width=0.9\linewidth]{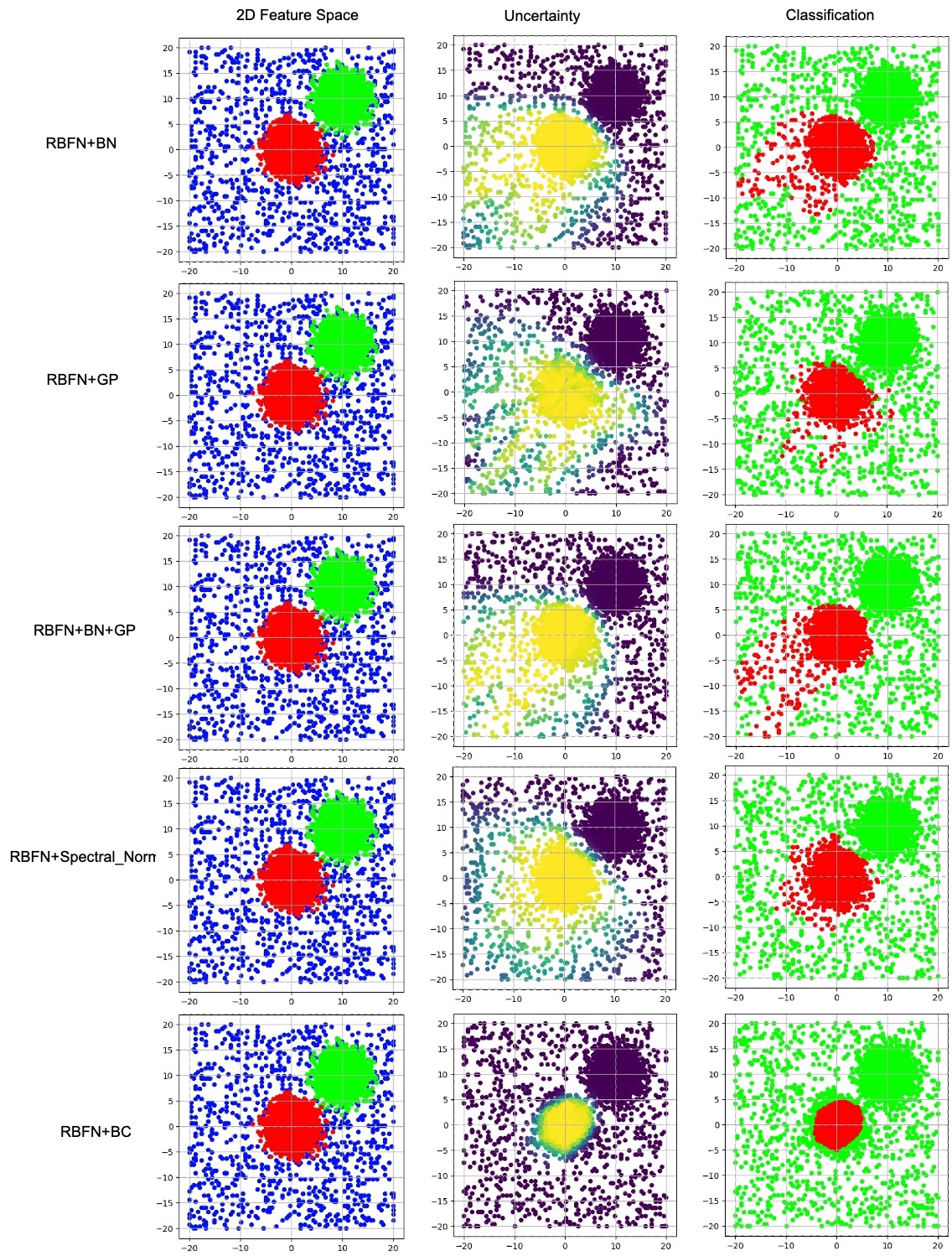}
\end{center}
 \caption{Visualization of the toy example with gaussian blobs on 2d space. Columns from left to right are the test data, uncertainty estimation results (brighter color means lower uncertainty) and classification into the center blob based on uncertainty. For each row from top to bottom, are RBFN+BatchNorm(BN), RBFN+GradientPenalty(GP), RBFN+BatchNorm+GradientPenalty, RBFN+SpectralNorm and RBFN+BoudnaryConstraint.}
\label{fig:gaussian_blobs}
\end{figure*}

\begin{figure*}[ht]
\begin{center}
\includegraphics[width=0.8\linewidth]{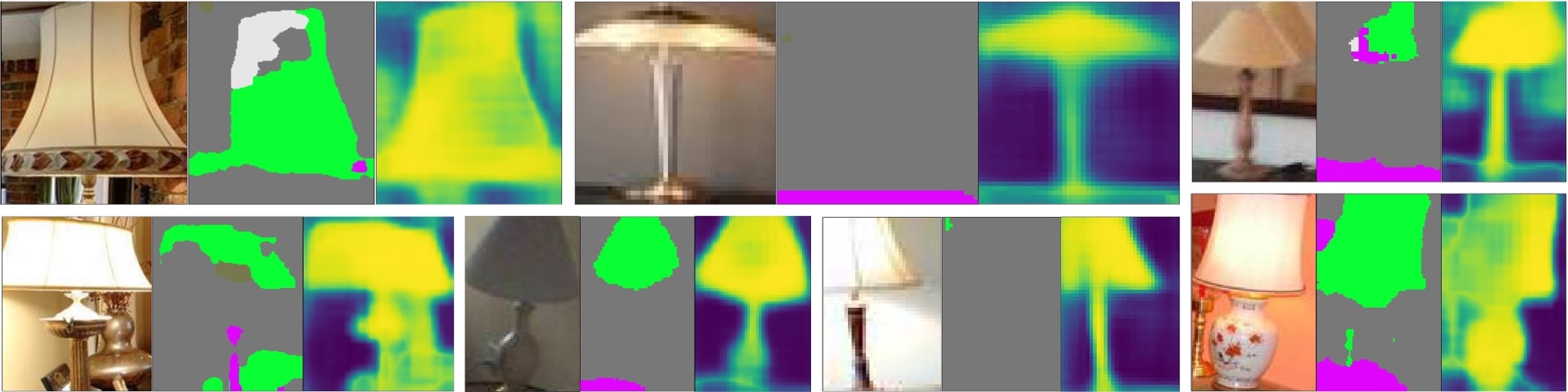}
\end{center}
%\vspace{-0.5cm}
 \caption{Proposal segmentation on lamps (unknown object) from ADE20K Dataset. For each tuple, we have the input proposal, semantic segmentation results and the uncertainty estimation.}
%\vspace{-1.5em}
\label{fig:ade20k}
\end{figure*}

\begin{figure*}[ht]
\begin{center}
\includegraphics[width=0.8\linewidth]{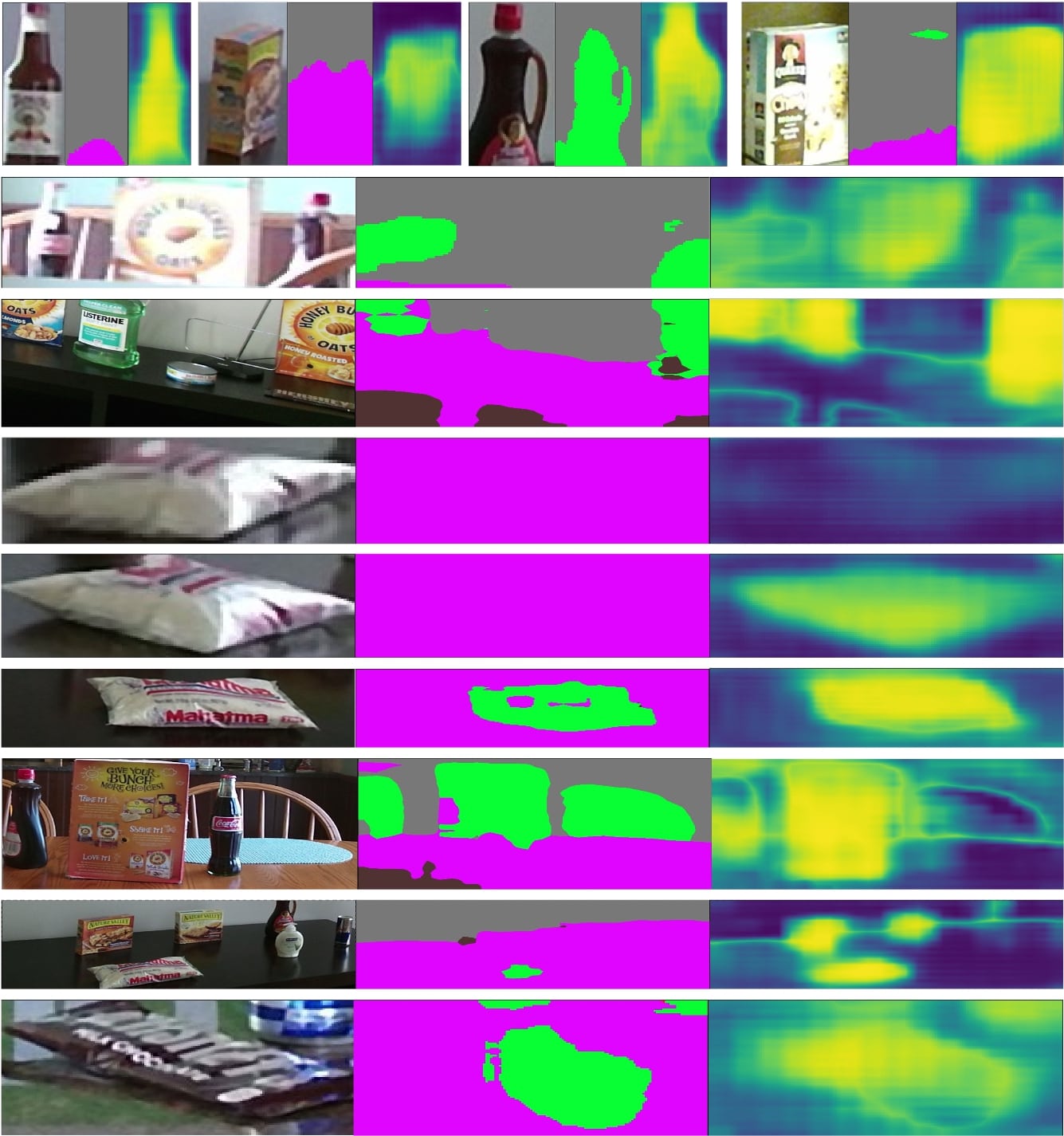}
\end{center}
%\vspace{-0.5cm}
 \caption{Proposal segmentation on novel instances from AVD Dataset.}
%\vspace{-1.5em}
\label{fig:avd}
\end{figure*}

We showed additional results of proposal segmentation (Fig.~\ref{fig:prop_seg}), whole image segmentation (Fig.~\ref{fig:whole_LF},~\ref{fig:whole_RA}).
The whole gaussian blob toy example is visualized at Fig~\ref{fig:gaussian_blobs}. More proposal segmentation results on indoor scenes are visualized at Fig~\ref{fig:ade20k},~\ref{fig:avd}.

\begin{figure*}[ht]
\begin{center}
\includegraphics[width=0.8\linewidth]{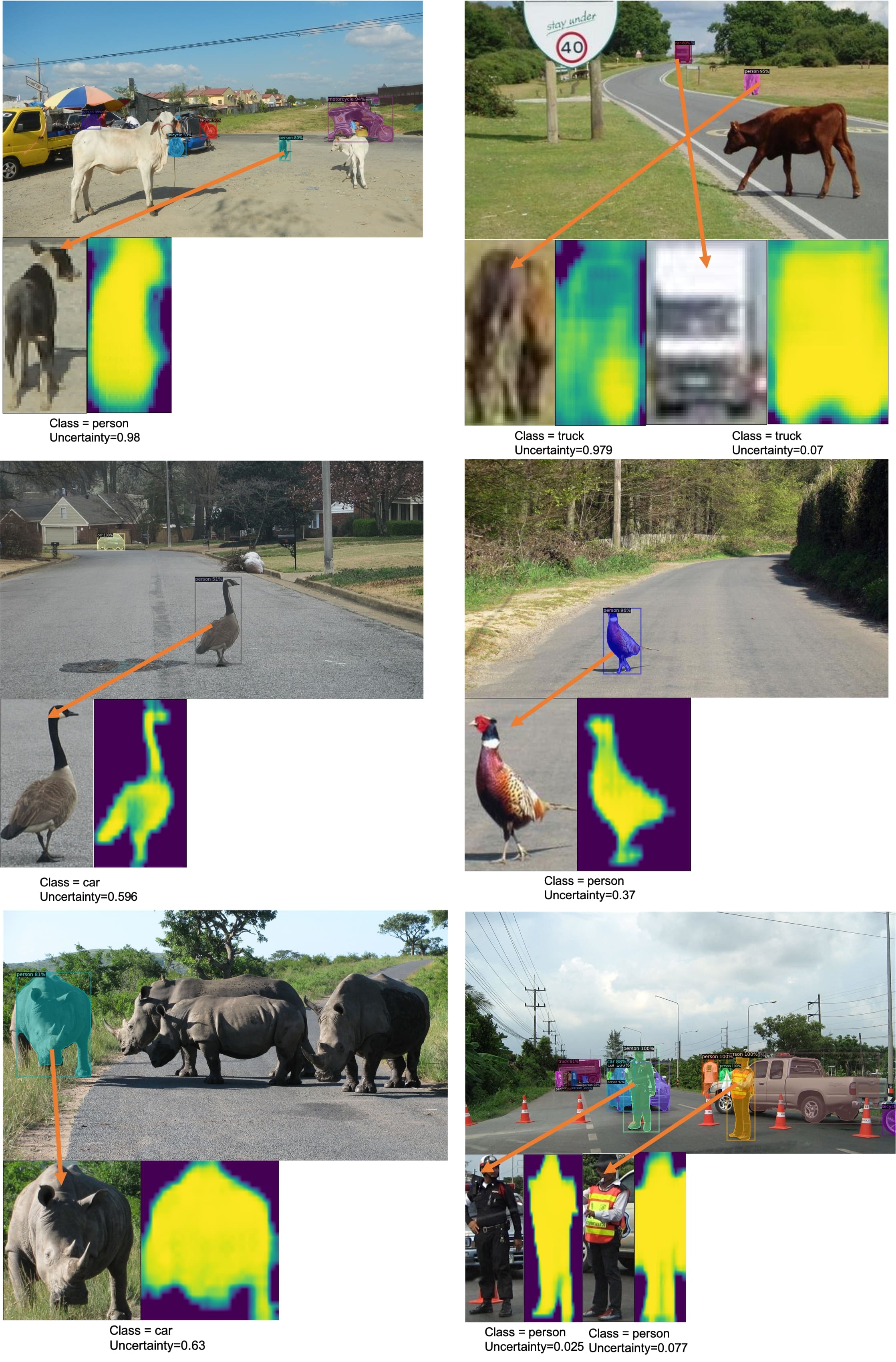}
\end{center}
 \caption{Compare Mask-RCNN predictions with our model.}
\label{fig:maskrcnn_faults}
\end{figure*}

\subsection{Model Architecture}
\begin{figure*}[ht]
\begin{center}
\includegraphics[width=0.8\linewidth]{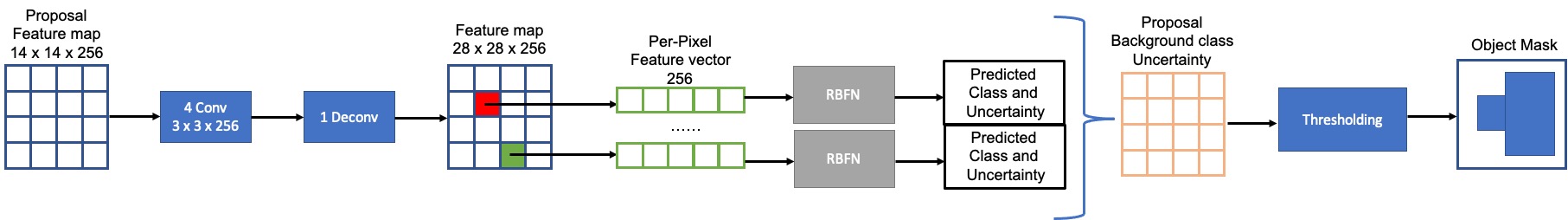}
\end{center}
   \caption{Proposal Segmentation Model}
\label{fig:prop_seg_arch}
\end{figure*}

\begin{figure*}[ht]
\begin{center}
\includegraphics[width=0.8\linewidth]{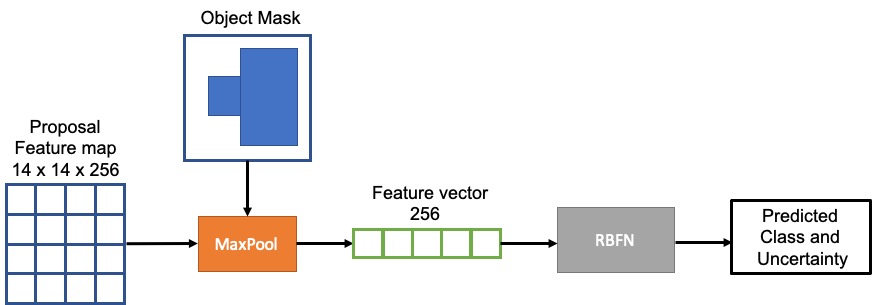}
\end{center}
   \caption{Proposal Classification Model}
\label{fig:prop_cls_arch}
\end{figure*}

\begin{figure*}[ht]
\begin{center}
\includegraphics[width=0.8\linewidth]{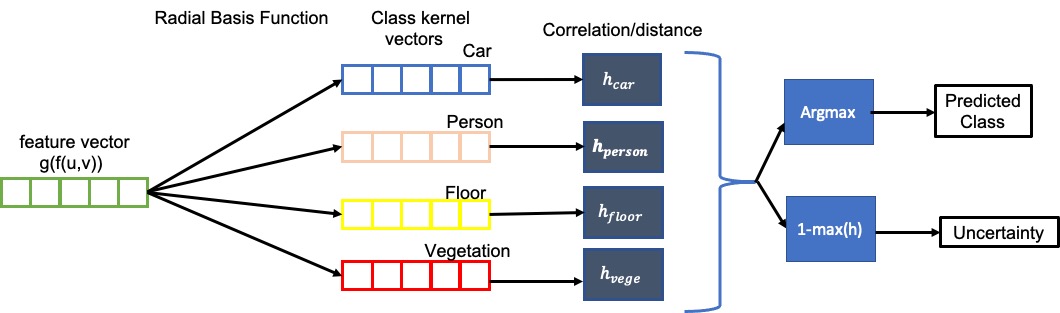}
\end{center}
   \caption{Radial Basis Function Network}
\label{fig:rbfn_arch}
\end{figure*}

Here we draw the figures of the architecture of the proposed proposal segmentation (Fig.~\ref{fig:prop_seg_arch}) and proposal classification (Fig.~\ref{fig:prop_cls_arch}) model. Details of RBF network is shown in Fig.~\ref{fig:rbfn_arch}.

\end{document}